\documentclass[fleqn,twoside]{article}
\usepackage{espcrc1}

\newcommand{\AmS}{{\protect\the\textfont2
  A\kern-.1667em\lower.5ex\hbox{M}\kern-.125emS}}
\title{Towards Bin Packing (preliminary problem survey, models with multiset estimates)
\thanks{
 The work is partially supported by The Russian Foundation for
 Basic Research, Grant No. 16-07-00092.}
}

\author{Mark Sh. Levin
\address{
 Inst. for Information Transmission Problems, Russian Academy of
 Sciences\\
 19 Bolshoj Karetny Lane, Moscow 127994, Russia\\
 E-mail: mslevin@acm.org
 }
 }

\begin{document}

\maketitle

\begin{abstract}
 The paper described a generalized integrated glance
 to bin packing problems including
 a brief literature survey
 and some new problem formulations for the cases of
 multiset estimates of items.
 A new systemic viewpoint to bin packing problems is suggested:
 (a) basic element sets
 (item set, bin set, item subset assigned to bin),
 (b) binary relation over the sets:
 relation over item set as compatibility,
 precedence, dominance;
 relation over items and bins
 (i.e., correspondence of items to bins).
%
%
 A special attention is targeted
 to the following versions of bin packing problems:
 (a) problem with multiset estimates of items,
 (b) problem with colored items
 (and some close problems).
%
 Applied examples of bin packing problems
 are considered:
 (i) planning in paper industry (framework of combinatorial problems),
 (ii) selection of information messages,
 (iii) packing of messages/information packages in WiMAX communication system
 (brief description).

~~~~~~~~~~~


\vspace{1pc}

\end{abstract}

\tableofcontents

\newcounter{cms}
\setlength{\unitlength}{1mm}

\newpage
\section{Introduction}

 Bin-packing problem is one of the well-known basic
  combinatorial optimization problems
 (e.g., \cite{brown71,cormen01,gar79,mar90}).
 The problem is a special case of
 one-dimensional ``cutting-stock'' problem \cite{gil63,val02}
 and the ``assembly-line balancing'' problem \cite{conw67}.
 Fig. 1 illustrates the relationship of one-dimensional bin-packing problems
 and some other combinatorial optimization problems.

\begin{center}
\begin{picture}(114,72)
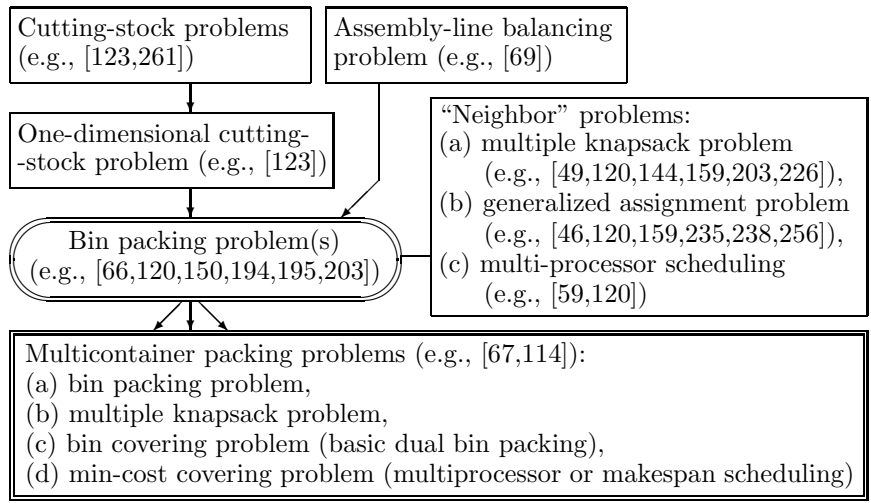
\put(16,00){\makebox(0,0)[bl]{Fig. 1. Bin-packing
 problems and their relationship}}

\put(00,61){\line(1,0){40}} \put(00,71){\line(1,0){40}}
\put(00,61){\line(0,1){10}} \put(40,61){\line(0,1){10}}

\put(01,66.5){\makebox(0,0)[bl]{Cutting-stock problems }}
\put(01,62.5){\makebox(0,0)[bl]{(e.g., \cite{gil63,val02}) }}

\put(24,61){\vector(0,-1){4}}


\put(00,47){\line(1,0){43}} \put(00,57){\line(1,0){43}}
\put(00,47){\line(0,1){10}} \put(43,47){\line(0,1){10}}

\put(01,52.5){\makebox(0,0)[bl]{One-dimensional cutting-}}
\put(01,48.5){\makebox(0,0)[bl]{-stock problem
(e.g., \cite{gil63}) }}

\put(24,47){\vector(0,-1){4}}

\put(42,61){\line(1,0){40}} \put(42,71){\line(1,0){40}}
\put(42,61){\line(0,1){10}} \put(82,61){\line(0,1){10}}

\put(43,66.5){\makebox(0,0)[bl]{Assembly-line balancing}}
\put(43,62.5){\makebox(0,0)[bl]{problem (e.g., \cite{conw67})}}


\put(49,61){\line(0,-1){13}} \put(49,48){\vector(-1,-1){05}}


\put(26,37.5){\oval(52,11)} \put(26,37.5){\oval(51,10)}

\put(07.8,38){\makebox(0,0)[bl]{Bin packing problem(s)}}

\put(02.5,34){\makebox(0,0)[bl]{(e.g.,
 \cite{cof02,gar79,john73,lodi02,lodi02a,mar90})}}

\put(56,30){\line(1,0){58}} \put(56,59){\line(1,0){58}}
\put(56,30){\line(0,1){29}} \put(114,30){\line(0,1){29}}

\put(57,55){\makebox(0,0)[bl]{``Neighbor'' problems:}}

\put(57,51){\makebox(0,0)[bl]{(a) multiple knapsack problem}}
\put(63,47){\makebox(0,0)[bl]{(e.g.,
 \cite{chek05,gar79,hung78,keller04,mar90,pisinger99}),}}

\put(57,43){\makebox(0,0)[bl]{(b) generalized assignment problem}}
\put(63,39){\makebox(0,0)[bl]{(e.g.,
  \cite{catt92,gar79,keller04,ross75,savel97,trick92}), }}

\put(57,35){\makebox(0,0)[bl]{(c) multi-processor scheduling}}

\put(63,31){\makebox(0,0)[bl]{(e.g., \cite{gar87,gar79}) }}

\put(56,38){\line(-1,0){4}}

\put(23,32){\vector(-1,-1){4}} \put(24,32){\vector(0,-1){4}}
\put(25,32){\vector(1,-1){4}}

\put(00,05.5){\line(1,0){114}} \put(00,28){\line(1,0){114}}
\put(00,05.5){\line(0,1){22.5}} \put(114,05.5){\line(0,1){22.5}}

\put(0.5,06){\line(1,0){113}} \put(0.5,27.5){\line(1,0){113}}
\put(0.5,06){\line(0,1){21.5}} \put(113.5,06){\line(0,1){21.5}}

\put(02,23){\makebox(0,0)[bl]{Multicontainer packing problems
  (e.g., \cite{cof07,fuk07}):}}

\put(02,19){\makebox(0,0)[bl]{(a) bin packing problem,}}

\put(02,15){\makebox(0,0)[bl]{(b) multiple knapsack problem,}}

\put(02,11){\makebox(0,0)[bl]{(c) bin covering problem (basic dual
bin packing),}}

\put(02,07){\makebox(0,0)[bl]{(d) min-cost covering problem
(multiprocessor or makespan scheduling)}}

\end{picture}
\end{center}

 The bin packing problem can be described as follows (Fig. 2).
 Initial information involves the following:
 ~(i) a set of items \(A = \{a_{1},...,a_{i},...,a_{n} \}\),
 each item \(a_{i}\) has a weight \(w_{i} \in (0,1]\);
%
 ~(ii) a set of bins (or one-dimensional containers, blocks)
  \(B = \{ B_{1},...,B_{\kappa},...,B_{m}\}\),
  capacity of each bin
   \(B_{\kappa}\) equals \(1\).
 The basic (classical)
  bin packing problem is
 (e.g., \cite{john73,john74a,john74,ull71}):


 Find a partition of the items such that:
 ~(a) each part of the item set is
 packed into the same bin while taking into account the bin capacity constraint
 (i.e., the sum of packed items in each bin  \(\leq 1\)),
 ~(b) the total number of used bins is minimized.


 This problem is one of basic NP-hard combinatorial optimization problems
 (e.g., \cite{gar79,karp72}).
  Fig. 2 illustrates the classic bin packing
  (i.e., packing the items into the minimal number of bins):
  \(6\) items are packed into \(3\) bins.

\begin{center}
\begin{picture}(110,60)
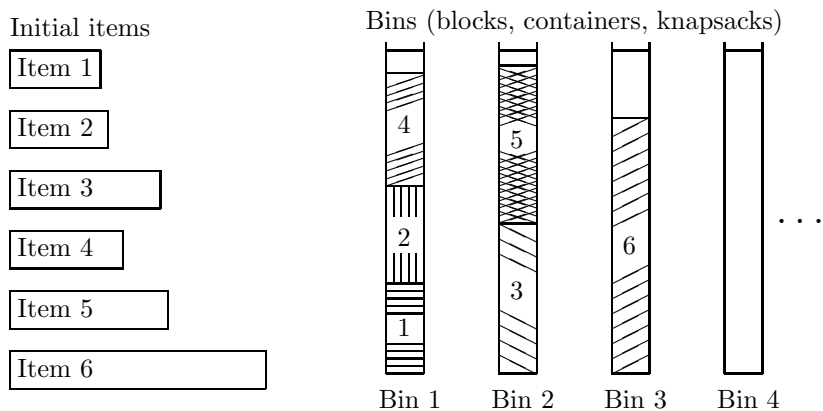
\put(18,00){\makebox(0,0)[bl]{Fig. 2. Illustration for classical
 bin-packing}}

\put(00,55){\makebox(0,0)[bl]{Initial items}}

\put(00,48){\line(1,0){12}} \put(00,53){\line(1,0){12}}
\put(00,48){\line(0,1){05}} \put(12,48){\line(0,1){05}}

\put(01,49.6){\makebox(0,0)[bl]{Item 1}}

\put(00,40){\line(1,0){13}} \put(00,45){\line(1,0){13}}
\put(00,40){\line(0,1){05}} \put(13,40){\line(0,1){05}}

\put(01,41.6){\makebox(0,0)[bl]{Item 2}}

\put(00,32){\line(1,0){20}} \put(00,37){\line(1,0){20}}
\put(00,32){\line(0,1){05}} \put(20,32){\line(0,1){05}}

\put(01,33.6){\makebox(0,0)[bl]{Item 3}}

\put(00,24){\line(1,0){15}} \put(00,29){\line(1,0){15}}
\put(00,24){\line(0,1){05}} \put(15,24){\line(0,1){05}}

\put(01,25.6){\makebox(0,0)[bl]{Item 4}}

\put(00,16){\line(1,0){21}} \put(00,21){\line(1,0){21}}
\put(00,16){\line(0,1){05}} \put(21,16){\line(0,1){05}}

\put(01,17.6){\makebox(0,0)[bl]{Item 5}}

\put(00,08){\line(1,0){34}} \put(00,13){\line(1,0){34}}
\put(00,08){\line(0,1){05}} \put(34,08){\line(0,1){05}}

\put(01,09.6){\makebox(0,0)[bl]{Item 6}}

\put(47.3,55){\makebox(0,0)[bl]{Bins (blocks, containers,
 knapsacks)}}

\put(50,10){\line(1,0){05}} \put(50,53){\line(1,0){05}}
\put(50,10){\line(0,1){44}} \put(55,10){\line(0,1){44}}

\put(49,05.5){\makebox(0,0)[bl]{Bin 1}}

\put(50,22){\line(1,0){05}} \put(51.5,15){\makebox(0,0)[bl]{1}}

\put(50,11){\line(1,0){05}} \put(50,12){\line(1,0){05}}
\put(50,13){\line(1,0){05}} \put(50,14){\line(1,0){05}}

\put(50,18){\line(1,0){05}} \put(50,19){\line(1,0){05}}
\put(50,20){\line(1,0){05}} \put(50,21){\line(1,0){05}}

\put(50,35){\line(1,0){05}} \put(51.5,27){\makebox(0,0)[bl]{2}}

\put(51,22){\line(0,1){04}} \put(52,22){\line(0,1){04}}
\put(53,22){\line(0,1){04}} \put(54,22){\line(0,1){04}}

\put(51,31){\line(0,1){04}} \put(52,31){\line(0,1){04}}
\put(53,31){\line(0,1){04}} \put(54,31){\line(0,1){04}}

\put(50,50){\line(1,0){05}} \put(51.5,42){\makebox(0,0)[bl]{4}}

\put(50,35){\line(3,1){05}} \put(50,36){\line(3,1){05}}
\put(50,37){\line(3,1){05}} \put(50,38){\line(3,1){05}}
\put(50,39){\line(3,1){05}}

\put(50,45){\line(3,1){05}} \put(50,46){\line(3,1){05}}
\put(50,47){\line(3,1){05}} \put(50,48){\line(3,1){05}}

\put(65,10){\line(1,0){05}} \put(65,53){\line(1,0){05}}
\put(65,10){\line(0,1){44}} \put(70,10){\line(0,1){44}}

\put(64,05.5){\makebox(0,0)[bl]{Bin 2}}

\put(65,30){\line(1,0){05}} \put(66.5,20){\makebox(0,0)[bl]{3}}


\put(65,30){\line(2,-1){05}} \put(65,28){\line(2,-1){05}}
\put(65,26){\line(2,-1){05}}

\put(70,16){\line(-2,1){05}} \put(70,14){\line(-2,1){05}}
\put(70,10){\line(-2,1){05}} \put(70,12){\line(-2,1){05}}

\put(65,51){\line(1,0){05}} \put(66.5,40){\makebox(0,0)[bl]{5}}

\put(65,49){\line(3,1){05}} \put(65,48){\line(3,1){05}}
\put(65,47){\line(3,1){05}} \put(65,46){\line(3,1){05}}
\put(65,45){\line(3,1){05}} \put(65,44){\line(3,1){05}}
\put(65,43){\line(3,1){05}}

\put(70,49){\line(-3,1){05}} \put(70,48){\line(-3,1){05}}
\put(70,47){\line(-3,1){05}} \put(70,46){\line(-3,1){05}}
\put(70,45){\line(-3,1){05}} \put(70,44){\line(-3,1){05}}
\put(70,43){\line(-3,1){05}}

\put(65,38){\line(3,1){05}} \put(65,37){\line(3,1){05}}
\put(65,36){\line(3,1){05}} \put(65,35){\line(3,1){05}}
\put(65,34){\line(3,1){05}} \put(65,33){\line(3,1){05}}
\put(65,32){\line(3,1){05}} \put(65,31){\line(3,1){05}}
\put(65,30){\line(3,1){05}}

\put(70,38){\line(-3,1){05}} \put(70,37){\line(-3,1){05}}
\put(70,36){\line(-3,1){05}} \put(70,35){\line(-3,1){05}}
\put(70,34){\line(-3,1){05}} \put(70,33){\line(-3,1){05}}
\put(70,32){\line(-3,1){05}} \put(70,31){\line(-3,1){05}}
\put(70,30){\line(-3,1){05}}

\put(80,10){\line(1,0){05}} \put(80,53){\line(1,0){05}}
\put(80,10){\line(0,1){44}} \put(85,10){\line(0,1){44}}

\put(79,05.5){\makebox(0,0)[bl]{Bin 3}}

\put(80,44){\line(1,0){05}}

\put(81.5,26){\makebox(0,0)[bl]{6}}

\put(80,10){\line(2,1){05}} \put(80,12){\line(2,1){05}}
\put(80,14){\line(2,1){05}} \put(80,16){\line(2,1){05}}
\put(80,18){\line(2,1){05}} \put(80,20){\line(2,1){05}}
\put(80,22){\line(2,1){05}}

\put(80,29.5){\line(2,1){05}} \put(80,31.5){\line(2,1){05}}
\put(80,33.5){\line(2,1){05}}  \put(80,35.5){\line(2,1){05}}
\put(80,37.5){\line(2,1){05}} \put(80,39.5){\line(2,1){05}}

 \put(80,41.5){\line(2,1){05}}

\put(95,10){\line(1,0){05}} \put(95,53){\line(1,0){05}}
\put(95,10){\line(0,1){44}} \put(100,10){\line(0,1){44}}

\put(94,05.5){\makebox(0,0)[bl]{Bin 4}}

\put(102,30){\makebox(0,0)[bl]{{\bf .~.~.}}}


\end{picture}
\end{center}

 Note the following basic types of items are examined
 (e.g., \cite{bansal09,bennell09,bennell10,bir05,cof07,dyck90,dyck92,fur86,galam94,gar79,lodi99b,lodi04,mar90,mar00,pis05,seid03,was07}):
 rectangular items,
 2D items,
 irregular shape items,
 variable sizes items,
 composite 2D items
 (including items with common components),
 3D items,
 multidimensional items,
 items as cylinders,
 items as circles, etc.
%
 A generalized illustration for bin packing problem
 is depicted in Fig. 3.

\begin{center}
\begin{picture}(109,57)

\put(09,00){\makebox(0,0)[bl]{Fig. 3. Generalized illustration
 for bin packing problem}}

\put(11,53){\makebox(0,0)[bl]{Some illustrative examples of items }}

\put(00,42){\line(1,0){06}}   \put(00,52){\line(1,0){06}}
\put(00,42){\line(0,1){10}}   \put(06,42){\line(0,1){10}}

\put(00,51){\line(1,0){06}}
\put(00,49){\line(1,0){06}} \put(00,50){\line(1,0){06}}
\put(00,47){\line(1,0){06}} \put(00,48){\line(1,0){06}}
\put(00,45){\line(1,0){06}} \put(00,46){\line(1,0){06}}
\put(00,43){\line(1,0){06}} \put(00,44){\line(1,0){06}}

\put(20,47){\oval(07,7)}

\put(32,42){\line(1,0){05}}   \put(32,47){\line(1,0){05}}
\put(32,42){\line(0,1){05}}   \put(37,42){\line(0,1){05}}

\put(32,47){\line(1,1){05}}   \put(37,47){\line(1,1){05}}
\put(37,42){\line(1,1){05}}

\put(42,47){\line(0,1){05}}
\put(37,52){\line(1,0){05}}

\put(51,42){\line(1,0){13}}   \put(51,51){\line(1,0){13}}
\put(51,42){\line(0,1){09}}   \put(64,42){\line(0,1){09}}

\put(51,46){\line(1,0){04}}   \put(55,46){\line(1,1){05}}
\put(60,51){\line(1,-1){04}}

\put(52,42){\line(0,1){04}} \put(53,42){\line(0,1){04}}
\put(54,42){\line(0,1){04}} \put(55,42){\line(0,1){04}}
\put(56,42){\line(0,1){05.2}} \put(57,42){\line(0,1){06.2}}
\put(58,42){\line(0,1){07.2}} \put(59,42){\line(0,1){08.2}}
\put(60,42){\line(0,1){09}} \put(61,42){\line(0,1){08.1}}
\put(62,42){\line(0,1){07.1}} \put(63,42){\line(0,1){06.1}}

\put(74,42){\line(1,0){15}} \put(74,57){\line(1,0){15}}
\put(74,42){\line(0,1){15}}   \put(89,42){\line(0,1){15}}

\put(74,47){\line(1,0){05}}   \put(79,42){\line(0,1){05}}

\put(74,42){\line(1,0){05}} \put(74,43){\line(1,0){05}}

\put(74,44){\line(1,0){05}} \put(74,45){\line(1,0){05}}
\put(74,46){\line(1,0){05}}

\put(79,47){\line(1,0){04}}   \put(79,51){\line(1,0){04}}
\put(79,47){\line(0,1){04}}   \put(83,47){\line(0,1){04}}

\put(80,47){\line(0,1){04}} \put(81,47){\line(0,1){04}}
\put(82,47){\line(0,1){04}}

\put(83,51){\line(1,0){06}}   \put(83,57){\line(1,0){06}}
\put(83,51){\line(0,1){06}}   \put(89,51){\line(0,1){06}}

\put(84,51){\line(0,1){06}} \put(85,51){\line(0,1){06}}
\put(86,51){\line(0,1){06}} \put(87,51){\line(0,1){06}}
\put(88,51){\line(0,1){06}}

\put(83,52){\line(1,0){06}} \put(83,53){\line(1,0){06}}
\put(83,54){\line(1,0){06}} \put(83,55){\line(1,0){06}}
\put(83,56){\line(1,0){06}}


\put(25,36.5){\makebox(0,0)[bl]{Set of items \(A = \{a_{1},a_{2},...,a_{n} \}\)}}

\put(20,29){\line(1,0){59}}   \put(20,35){\line(1,0){59}}
\put(20,29){\line(0,1){06}}   \put(79,29){\line(0,1){06}}
\put(20.5,29){\line(0,1){06}} \put(78.5,29){\line(0,1){06}}

\put(26.5,30.7){\makebox(0,0)[bl]{Assignment of items into bins}}

\put(25,29){\vector(-1,-1){4}}
\put(37,29){\vector(0,-1){4}}
\put(49.5,29){\vector(0,-1){4}}
\put(62,29){\vector(0,-1){4}}
\put(74,29){\vector(1,-1){4}}


\put(00,10){\line(1,0){28}}
\put(00,10){\line(0,1){16}} \put(28,10){\line(0,1){16}}

\put(00,24){\line(1,0){04}} \put(05.5,24){\line(1,0){04}}
\put(11,24){\line(1,0){04}} \put(16.5,24){\line(1,0){04}}
\put(22,24){\line(1,0){04}}

\put(08,05.8){\makebox(0,0)[bl]{Bin \(B_{1}\) }}

\put(01,15){\makebox(0,0)[bl]{\(A^{1} = \{a^{1}_{1},...,a^{1}_{q_{1}}\}\) }}

\put(02.7,11){\makebox(0,0)[bl]{(packed items)}}


\put(30,17){\makebox(0,0)[bl]{{\bf ... }}}

\put(35,10){\line(1,0){29}}
\put(35,10){\line(0,1){13}} \put(64,10){\line(0,1){13}}

\put(35,21){\line(1,0){04}} \put(40,21){\line(1,0){04}}
\put(45,21){\line(1,0){04}} \put(50,21){\line(1,0){04}}
\put(55,21){\line(1,0){04}} \put(60,21){\line(1,0){04}}

\put(45,5.8){\makebox(0,0)[bl]{Bin \(B_{\kappa}\) }}

\put(36,15){\makebox(0,0)[bl]{\(A^{\kappa} = \{a^{\kappa}_{1},...,a^{\kappa}_{q_{\kappa}}\}\) }}

\put(38.5,11){\makebox(0,0)[bl]{(packed items)}}


\put(66,17){\makebox(0,0)[bl]{{\bf ... }}}

\put(71,10){\line(1,0){32}}
\put(71,10){\line(0,1){15}} \put(103,10){\line(0,1){15}}

\put(71,23){\line(1,0){04}} \put(76,23){\line(1,0){04}}
\put(81,23){\line(1,0){04}} \put(86,23){\line(1,0){04}}
\put(91,23){\line(1,0){04}} \put(96,23){\line(1,0){04}}

\put(81,5.8){\makebox(0,0)[bl]{Bin \(B_{m}\) }}

\put(72,15){\makebox(0,0)[bl]{\(A^{m} = \{a^{m}_{1},...,a^{m}_{q_{m}}\}\) }}

\put(75,11){\makebox(0,0)[bl]{(packed items)}}


\put(105,17){\makebox(0,0)[bl]{{\bf ... }}}

\end{picture}
\end{center}

 Further,
 it is necessary to point out binary relations:


 {\bf I.} Binary relations over initial items and bins
 (items \(A = \{a_{1},a_{2},...,a_{n} \}\),
  bins \(B = \{B_{1},...,B_{\kappa},...,B_{m} \}\)):

 {\it 1.1.} correspondence of items to bins or preference
 (for each item)
 as binary relation (or weighted binary relation):~
 \(R_{A \times B}\).

 {\bf II.} Binary relation over items:

 {\it 2.1.} conflicts as a binary relations for item pairs that
 can not be assigned into the same bin:~
 \(R^{confl}_{A \times A}\)
 (this can be considered as a part of the next relation),

 {\it 2.2.} compatibility (e.g., by type/color) as binary relation for  items
 which are compatible (e.g., for assignment to the same bin,
 to be neighbor in the same bin):~
 \(R^{compt}_{L \times L}\),
 here a weighted binary relation can be useful
 (e.g., for colors, including non-symmetric binary relation for neighborhood),

 {\it 2.3.} compatibility (e.g., by common components, for multi-component items),
 close to previous case (this may be crucial for ``intersection'' of items):~
 \(R^{compt-com}_{A \times A}\),

 {\it 2.4.} precedence over items (this is important in the case of
 ordering of items which are assigned into the same bin):~
 \(R^{prec}_{A \times A}\),

  {\it 2.5.} importance (dominance, preference) of items from the viewpoint of the first
 assignment to bins,
 as a linear ordering
 or poset-like structure over items:~
  \(G(A,E^{dom})\)
  (the poset-like structure may be based on multicriteria estimates or multiset estimates  of items).

 {\bf III.} Binary relations over bins:

 {\it 3.1.} importance of bins from the viewpoint of the first usage,
 as a linear ordering
 or poset-like structure) over bins:~
  \(G(B,E^{imp})\)
 (the poset-like structure may be based on multicriteria estimates
 or multiset estimates
 of bins).


 Numerical examples of the above-mentioned relations
 are presented as follows
(on the basis of example from Fig. 2: six items and four bins):
 (i) correspondence of items to bins
  \(R_{A \times B}\)
  (Table 1),
 (ii)  relation of item conflict
 \(R^{confl}_{A \times A}\)
 (Table 2),
 (iii) relation of item compatibility
 \(R^{comp}_{A \times A}\)
 (e.g., by type/color)
  (Table 3),
 (iv) precedence relation over items \(R^{prec}_{A \times A}\)
 (Fig. 4 ),
 (v)  (relation of dominance over items \(G(A,E^{dom})\)
   (Fig. 5),
 and
 (vi) relation of importance over bins \(G(B,E^{imp})\)
 (Fig. 6).

 Further, the solution of the bin packing problem can be examined as
 the following (i.e., assignment of items into bins, a Boolean matrix):
 \(S = \{A^{1},...,A^{\kappa},...,A^{k},...,A^{m}  \}\)
 where
 \( | A^{\kappa_{1}} \bigcap  A^{\kappa_{2}} | = 0\)
  ~\(\forall \kappa_{1},\kappa_{2} = \overline{1,m} \)
 (i.e., the intersection is empty),
 \( A = \bigcup_{\kappa=1}^{m} A^{\kappa} \).

 For classic bin packing problem
 (i.e., minimization of used bins),
 \(| A^{\kappa}| = 0\) \(\forall \kappa = \overline{k+1,m}\)
 (the first \(k\) bins are used)
 and  \( A = \bigcup_{\kappa=1}^{m} A^{\kappa} \).

 In inverse bin packing problem
 (maximization of assigned items into the limited number of bins),
 a part of the most important items are assigned into
 \(m\) bins:~
  \( \bigcup_{\kappa=1}^{m} A^{\kappa} \subseteq A \).

 Additional requirements to packing solutions are the following
 (i.e., fulfilment of the constraints):

 {\bf 1. Correspondence of item to bin.}
 The following has to be satisfied:
 ~ \( a_{i} \in A_{\kappa}\) {\bf If }
 \( (a_{i},B_{\kappa}) \in R_{A \times B}\).

 {\bf 2. Importance/dominance of items.} This corresponds to inverse problem:
 ~ {\bf If} \( (a_{i_{1}},a_{i_{2}}) \in R^{dom}_{A \times A}\)
 (i.e., \( a_{i_{1}} \succeq a_{i_{2}} \)
 {\bf Then} three cases are correct:
 (a) both \(a_{i_{1}}\) and \(a_{i_{2}}\) are assigned to  bin(s),
 (b) both \(a_{i_{1}}\) and \(a_{i_{2}}\) are not assigned to  bin(s),
 (c) \(a_{i_{1}}\) is assigned to bin and \(a_{i_{2}}\) is not assigned to  bin.

 {\bf 3. Item precedence.}
 In the case of precedence constraint(s) according
 the above-mentioned precedence relations over items
 \(R^{prec}_{A \times A}\),
  the items have to be linear ordered in each bin
 (for each bin, i.e., \(\forall \kappa \)):
 ~~~{\bf If} \( (a_{i_{1}},a_{i_{2}}) \in R^{prec}_{A \times A}\)
 and \( a_{i_{1}},a_{i_{2}} \in A_{\kappa}\)
  {\bf Then} \( a_{i_{1}} \rightarrow a_{i_{2}}\).

 {\bf 4. Item conflicts.} In the case of conflict constraints,
 the following has to be satisfied:

 \( a_{i_{1}},a_{i_{2}} \in A_{\kappa}\) {\bf If }
 \( (a_{i_{1}},a_{i_{2}}) \overline{\in} R^{confl}_{A \times A}\).

 {\bf 5. Item compatibility.} In the case of compatibility constraints,
 the following has to be satisfied:

 \( a_{i_{1}},a_{i_{2}} \in A_{\kappa}\) {\bf If }
 \( (a_{i_{1}},a_{i_{2}}) \in R^{comp}_{A \times A}\).

 In general, it is possible to use some penalty functions
 in the cases when the constraints are not satisfied.

\begin{center}
{\bf Table 1.}  Correspondence of items to bins   \(R_{L \times B}\) \\
\begin{tabular}{| c | l|l|l| l |}
\hline
 Item \(a_{i}\) \(\backslash\) bin \(B_{\kappa}\)  &\(B_{1}\)&\(B_{2}\)&\(B_{3}\)&\(B_{4}\)  \\
\hline
 \(a_{1}\)&  \(3\) &  \(2\)&  \(1\)&  \(0\) \\
 \(a_{2}\)&  \(3\) &  \(1\)&  \(0\)&  \(0\) \\
 \(a_{3}\)&  \(1\) &  \(3\)&  \(2\)&  \(0\) \\
 \(a_{4}\)&  \(3\) &  \(2\)&  \(2\)&  \(0\) \\
 \(a_{5}\)&  \(1\) &  \(3\)&  \(1\)&  \(1\) \\
 \(a_{6}\)&  \(2\) &  \(3\)&  \(3\)&  \(1\) \\

\hline
\end{tabular}
\end{center}

\begin{center}
 {\bf Table 2.}  Relation on item conflict \(R^{confl}_{A \times A}\) \\
\begin{tabular}{| c | l|l|l| l|l|l |}
\hline
 Item \(a_{i}\) \(\backslash\) item \(a_{j}\)  &\(a_{1}\)&\(a_{2}\)&\(a_{3}\)&\(a_{4}\) &\(a_{5}\)&\(a_{6}\)  \\
\hline
 \(a_{1}\)&  \(\star\)&\(1\)&\(1\)&\(1\)&\(0\)&\(0\)  \\
 \(a_{2}\)&  \(1\)&\(\star\)&\(1\)&\(1\)&\(1\)&\(0\)  \\
 \(a_{3}\)&  \(1\)&\(3\)&\(\star\)&\(4\)&\(1\)&\(0\) \\
 \(a_{4}\)&  \(1\)&\(1\)&\(1\)&\(\star\)&\(1\)&\(0\)  \\
 \(a_{5}\)&  \(1\)&\(1\)&\(1\)&\(1\)&\(\star\)&\(0\)  \\
 \(a_{6}\)&  \(1\)&\(0\)&\(0\)&\(0\)&\(0\)&\(\star\)  \\

\hline
\end{tabular}
\end{center}

\begin{center}
 {\bf Table 3.}  Relation on item compatibility
  \(R^{comp}_{A \times A}\) \\
\begin{tabular}{| c | l|l|l| l|l|l |}
\hline
 Item \(a_{i}\) \(\backslash\) item \(a_{j}\)  &\(a_{1}\)&\(a_{2}\)&\(a_{3}\)&\(a_{4}\) &\(a_{5}\)&\(a_{6}\)  \\
\hline
 \(a_{1}\)&  \(\star\)&\(1\)&\(1\)&\(1\)&\(0\)&\(0\)  \\
 \(a_{2}\)&  \(1\)&\(\star\)&\(1\)&\(1\)&\(1\)&\(0\)  \\
 \(a_{3}\)&  \(1\)&\(3\)&\(\star\)&\(4\)&\(1\)&\(0\) \\
 \(a_{4}\)&  \(1\)&\(1\)&\(1\)&\(\star\)&\(1\)&\(0\)  \\
 \(a_{5}\)&  \(1\)&\(1\)&\(1\)&\(1\)&\(\star\)&\(0\)  \\
 \(a_{6}\)&  \(1\)&\(0\)&\(0\)&\(0\)&\(0\)&\(\star\)  \\

\hline
\end{tabular}
\end{center}

\begin{center}
\begin{picture}(80,26)

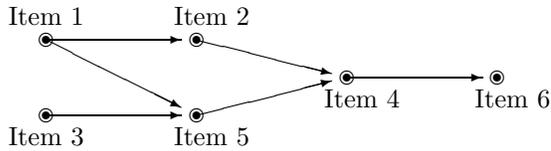
\put(07,00){\makebox(0,0)[bl]{Fig. 4. Precedence
 over items  \(R^{prec}_{A \times A}\)}}

\put(05,20){\circle*{1.1}} \put(05,20){\circle{1.9}}
\put(00,22){\makebox(0,0)[bl]{Item \(1\)}}

\put(05,20){\vector(1,0){18}} \put(05,20){\vector(2,-1){18}}

\put(25,20){\circle*{1.1}} \put(25,20){\circle{1.9}}
\put(22,22){\makebox(0,0)[bl]{Item \(2\)}}

\put(25,20){\vector(4,-1){18}}


\put(05,10){\circle*{1.1}} \put(05,10){\circle{1.9}}
\put(00,06){\makebox(0,0)[bl]{Item \(3\)}}

\put(05,10){\vector(1,0){18}}

\put(25,10){\circle*{1.1}} \put(25,10){\circle{1.9}}
\put(22,06){\makebox(0,0)[bl]{Item \(5\)}}

\put(25,10){\vector(4,1){18}}

\put(45,15){\circle*{1.1}} \put(45,15){\circle{1.9}}
\put(42,11){\makebox(0,0)[bl]{Item \(4\)}}

\put(45,15){\vector(1,0){18}}

\put(65,15){\circle*{1.1}} \put(65,15){\circle{1.9}}
\put(62,11){\makebox(0,0)[bl]{Item \(6\)}}

\end{picture}
%
\begin{picture}(50,26)

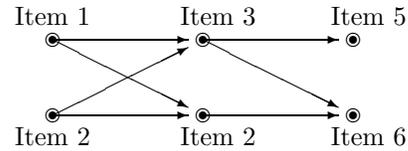
\put(00,00){\makebox(0,0)[bl]{Fig. 5. Importance of items \(G(A,E^{dom})\) }}

\put(05,20){\circle*{1.1}} \put(05,20){\circle{1.9}}
\put(00,22){\makebox(0,0)[bl]{Item \(1\)}}

\put(05,20){\vector(1,0){18}} \put(05,20){\vector(2,-1){18}}

\put(25,20){\vector(2,-1){18}}

\put(25,20){\circle*{1.1}} \put(25,20){\circle{1.9}}
\put(22,22){\makebox(0,0)[bl]{Item \(3\)}}

\put(25,20){\vector(1,0){18}}

\put(05,10){\vector(2,1){18}}


\put(05,10){\circle*{1.1}} \put(05,10){\circle{1.9}}
\put(00,06){\makebox(0,0)[bl]{Item \(2\)}}

\put(05,10){\vector(1,0){18}}

\put(25,10){\circle*{1.1}} \put(25,10){\circle{1.9}}
\put(22,06){\makebox(0,0)[bl]{Item \(2\)}}

\put(25,10){\vector(1,0){18}}

\put(45,20){\circle*{1.1}} \put(45,20){\circle{1.9}}
\put(42,22){\makebox(0,0)[bl]{Item \(5\)}}

\put(45,10){\circle*{1.1}} \put(45,10){\circle{1.9}}
\put(42,06){\makebox(0,0)[bl]{Item \(6\)}}

\end{picture}
\end{center}

\begin{center}
\begin{picture}(50,26)

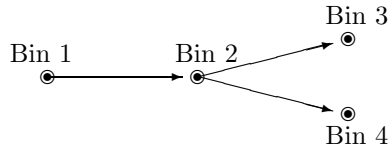
\put(00,00){\makebox(0,0)[bl]{Fig. 6. Importance of bins \(G(B,E^{imp})\) }}

\put(05,15){\circle*{1.1}} \put(05,15){\circle{1.9}}
\put(00,17){\makebox(0,0)[bl]{Bin \(1\)}}

\put(05,15){\vector(1,0){18}}

\put(25,15){\circle*{1.1}} \put(25,15){\circle{1.9}}
\put(22,17){\makebox(0,0)[bl]{Bin \(2\)}}

\put(25,15){\vector(4,1){18}}

\put(25,15){\vector(4,-1){18}}


\put(45,20){\circle*{1.1}} \put(45,20){\circle{1.9}}
\put(42,22){\makebox(0,0)[bl]{Bin \(3\)}}

\put(45,10){\circle*{1.1}} \put(45,10){\circle{1.9}}
\put(42,06){\makebox(0,0)[bl]{Bin \(4\)}}

\end{picture}
\end{center}

 Numerous publications have already addressed and analyzed
 various versions of static and dynamic bin packing problems.
  Many surveys on  BPPs have been published
 (e.g., \cite{bel03,coff00,cof02a,cof02,cof07,cof13,del15,fur16,gar79,lodi02a,mar90,swee92}).
%
%
 The basic taxonomies/typologies of bin packing problems have been examined
 in
 \cite{cof07,dyck90,dyck92,lodi99b,was07}.

 Some basic versions of bin packing problems (BPPs) are listed in Table 4
 and
  special classes of bin packing problems
 (e.g., with relations among items)
 are pointed out in Table 5.
 Table 6 contains a list of main applications
 of bin packing problems.

%
 Many surveys on algorithms for bin packing problems have been
 published
 (e.g.,
 \cite{bel03,coff97,cof02,cof13,del15,eps12,fuk07,hop97,hop01a,john73,mar90,reev96,tera10}).
 Basic algorithmic approaches are listed
 in Table 7, Table 8, and Table 9.


\begin{center}
{\bf Table 4.} Main bin packing problem formulations   \\
\begin{tabular}{| c | l| l |}
\hline
 No.  &Problem &Some source(s) \\
\hline

 I.&Basic bin packing problems:&\\

 1.1.&Classic one-dimensional bin-packing&\cite{john73,john74a,john74,ull71}\\
 1.2.&Bin-packing with discrete item sizes&\cite{cof97dis,coff00}\\
 1.3.& Linear programming formulation&\cite{val02}\\
 1.4.&Variable sized bin packing&\cite{boyar12,csirik89,fries86,fur86,kang03,murg87,seid01,seid03}\\
 1.5.&Maximum resource bin packing problem&\cite{boyar06}\\
 1.6.& Bin packing with cardinality (maximization, constraints)
 &\cite{bab04,labbe03,peet06}\\
 1.7.&Unrestrcited
 black and white bin packing&\cite{bal15}\\

 1.8.&Bin packing with rejection (including variable sized)&\cite{bein08,dar12,epstein06}\\

 1.9.&Bin packing with item fragmentation&\cite{casa14}\\

 1.10.& Bin packing games&\cite{kern12,kui98}\\

\hline
 II.&Multidimensional bin packing problems:&\\

 2.1.&2D bin-packing &\cite{alv09,char11,cui16,chung82,dar12,gar79,hop99,hop01}\\

    &  &\cite{hop01a,lodi99b,lodi02,lodi02a,lopez13,mar90,tera10}\\

 2.2.& Oriented 2D bin packing &\cite{lodi99a,lodi99b} \\

 2.3.&Othogonal 2D bin packing & \cite{bak80,leung03}\\

 2.4.&2D bin-packing with variable sizes (and costs)
 &\cite{chwa11,hong14,pis05}\\

 2.5.& 2D bin packing with due dates & \cite{bennell13}\\

 2.6.&2D bin-packing with guillotine constraints&\cite{char11,han13,krog95}\\

 2.7.&2D irregular shape bin packing&\cite{alb80,blaz93,han13}\\

 2.8.&
  3D bin packing &\cite{dai94,dow91,mack10,mar00,mar07,szy94}\\

 2.9.&Multi-dimensional bin-packing (vector packing) &\cite{bansal06,bansal09,csirik93,galam94,gar78,lodi04}\\

\hline
 III.&Online and dynamic bin packing problems:&\\

 3.1.&Online bin packing &\cite{bal15a,cof02a,eps06,gam00,seid02,seid03,van95}\\

 3.2.&Online bin packing with two item sizes
 &\cite{eps08a,gut06a}\\

 3.3.& Online bin packing with advise& \cite{boyar16}\\

 3.4.&Online variable-sized bin packing&\cite{csirik89,kin88,zhang97}\\

 3.5.&On-line bin-packing with cardinality constraints&\cite{bab04}\\


 3.6.&Dynamic bin packing problems& \cite{cof83}\\

 3.7.&Bin packing with controllable item sizes
 &\cite{corr08}\\

 3.8.&Batched bin packing&\cite{gutin05}\\

\hline
 IV.&
 Dual/inverse bin packing problems:   &\\

 4.1.&Inverse
  problem (maximizing the number of packed items,&\cite{ass84,bruno85,cof78,chung12,labbe95,labbe03,peet06}\\
  &maximum cardinality bin packing)&\\

 4.2.& Dual bin packing with items of random sizes &\cite{rhee93} \\

 4.3.&On-line dual bin packing& \cite{boyar01,csirik88}\\

 4.4.&On-line variable-sized dual problem&\cite{eps03}\\

 4.5.& ``Maximization'' of total preference estimate for& \cite{fur86}\\
     &packed items (preference relation over item set)&\\

 4.6.&Inverse
  bin packing with multiset estimates&This paper\\

\hline
  V.&Multi-stage bin packing problems:&\\
 5.1.&Three-stage two-dimensional bin packing &\cite{puch07}\\

 5.2.&Multi-stage bin packing&\\

\hline
 VI.& Bin packing problems in game theory perspective:&\\
 6.1.& Bin-packing of
 selfish items &\cite{bilo06,eps11,eps16,han13a,kleiman11,miyaz09,yu08}\\

 6.2.& Generalized selfish bin-packing &\cite{dosa12}\\


\hline
\end{tabular}
\end{center}

\newpage
\begin{center}
{\bf Table 5.} Bin packing problems with multi-component items, with
 binary relations \\
%
\begin{tabular}{| c | l| l |}
\hline
 No.  &Problem &Some source(s) \\

\hline
 I.&Problems with multi-component items/items fragmentation:&\\

 1.1.& Bin packing with multi-component items&\cite{fur86}\\

 1.2.&Packing with item fragmentation&\cite{shach08}\\

 1.3.&Packet scheduling with fragmentation&\cite{naam02} \\

\hline
 II.&Colored bin packing:&\\

 2.1.&Offline black and white bin packing&\cite{bal15}\\

 2.2.&Online black and white bin packing&\cite{bal15a}\\

 2.3.&Colored bin packing &\cite{chungf06,twigl15}\\

 2.4.& Offline colored bin packing&\cite{twigl15}\\

 2.5.&Online colored bin packing&\cite{bohm14,twigl15}\\

 2.6.&Online bin coloring (packing with minimum colors)&\cite{krumke01}\\


 2.7.&Composite planning framework in paper production system&This paper\\

\hline
 III.&Multcriteria/multobjective bin packing, relations over items:&\\

 3.1.&Bin packing  with conflicts&\cite{gend04,eps08,fern10,jans99,sad13}\\

 3.2.& Bin packing with multicriteria items&\cite{fur86}\\

 3.3.&Multi-objective bin packing &\cite{liud08,nad14}\\

 3.4.&Multi-objective bin packing with rotations
 &\cite{fern13}\\

 3.5.&Problems with preference over items&\cite{fur86}\\

 3.6.&Problems with precedence among items&\cite{del12,sch09,per16}\\

\hline
\end{tabular}
\end{center}

~~~

 A general classification scheme for bin packing problems has been
 suggested in
 \cite{cof07}:

~~

 {\bf arena} ~\(|\)~ {\bf objective function} ~\(|\)~ {\bf algorithm class} ~\(|\)~
 {\bf results} ~\(|\)~ {\bf constraints}

~~

 where the scheme components are as follows:
 (a) {\bf arena} describes types of bins (e.g., sizes, etc),
 (b) {\bf objective function} describes types of problem
 (i.e., minimum of bin, minimum of ``makespan'', etc.),
 (c) {\bf algorithm class} describes types of algorithm
 (e.g., offline, online, complexity estimate, greedy-type,
 etc.),
 (d)  {\bf constraints} describes quality of solution, e.g.,
  asymptotic worst case ratios,
  absolute worst case,
  average case, etc.,
 (e)  {\bf constraints} describes
 bounds on item sizes,
 bound on the number of items which can be packed in a bin,
 binary relation over item set
 (e.g., items \(a_{\iota_{1}}\) and \(a_{\iota_{2}}\)
 can not be
 put into the same bin), etc.

  Fig. 7 illustrates the basic trends in modifications
  of bin packing problems:
  (1) multicriteria (multi-objective) bin packing,
  (2) bin packing problems under uncertainty (e.g., fuzzy set
  usage of estimates),
  (3) examination of additional relations over items and over
  bins,
  (4) dynamic bin packing.

 This paper
 addresses the bin packing problem survey and
 some new formulations of bin packing problems:
 (a) with relations
  over item set,
 (b) with multiset estimates of items.

\newpage
\begin{center}
{\bf Table 6.} Some applications of bin packing problems \\
\begin{tabular}{| c | l| l |}
\hline
 No.  & Domain(s)/Problem(s)
  &Some source(s) \\

 \hline
 I.&Basic applications:&\\

 1.1.&Table formatting&\cite{john74}\\

 1.2.&Prepaging&\cite{john74}\\

 1.3.&File allocation, storage allocation&\cite{cof79,john74,ull71}\\

 1.4.&Processor allocation&\cite{cof79,jans99}\\

 1.5.&Mutli-processor scheduling &\cite{gar87,cof79,conw67,hub94,van95}\\

 1.6.&Examination timetabling & \cite{lap84}\\

 \hline
 II.&Industrial applications:&\\

 2.1.&Packing systems in industry &\cite{hop99,whe93}\\

 2.2.&Liquid loading problem  &\cite{chris79}\\

 2.3.&Assembly line balancing& \cite{per16}\\

 2.4.&Filling up containers&\cite{gar79}\\

 2.5.&Loading tracks with weights capacity constraints&\cite{gar79}\\

 2.6.&Vehicle container loading problem&\cite{eil71,hall89,jans99}\\

 2.7.&Loading of tractor trailer trucks& \cite{liud08}\\

 2.8.&Loading of cargo airplanes& \cite{liud08}\\

 2.9.&Loading of containers into ships &\cite{good06,imai01,kim04,liud08}\\

 2.10.&Packing in design automation&\cite{dai94,szy94}\\

 2.11.&Delivery problem & \cite{mar90}\\

 2.12.&Configuraiton of support tools for satellite mission
 &\cite{fur86}\\

\hline
 III.& Applications in distributed  computing:&\\

 3.1.&Assignment of processes to processors&\cite{jans99}\\

 3.2.&Allocating jobs in distributed
  computing systems (grids, etc.)&\cite{twigl15}\\

 3.3.&Data placement on parallel discs &\cite{gol00,kas06} \\

 3.4.&Dynamic resource allocation in cloud data centers
 &\cite{wolke15}\\


 3.5.&Periodic task scheduling in real-time distributional control systems
 & \cite{zhu11}\\

 & (e.g., automobile electronic control system, satellite control system, &\\
 &  medical equipment's electronic control system)&\\

\hline
 IV.&Applications in networking:&\\

 4.1.&Routing and wavelength assignment in optical networks &\cite{skor07}\\

 4.2.&Bandwidht allocation (e.g., channel assignment)&\cite{corr08}\\

 4.3.& Video-on-demand systems &\cite{xav08}\\

 4.4.&Creating file backups in media&\cite{gar79}\\

 4.5.&Allocating files in P2P networks &\cite{twigl15}\\

 4.6.&Packet scheduling with fragmentation&\cite{naam02} \\

 4.7.& Selection of messages/packages in communication system
   & This paper\\

 4.8.& 2D packing  for mobile WiMAX
 (e.g., data location

  &\cite{cic10,cic11,cic14,lodi11,mar14}\\

 &in IEEE.802.16/OFDMA)&\\

 4.9.&Resource allocation in multispot
 MFTDMA satellite networks
 &\cite{alouf06}\\

\hline
 V.&Some contemporary applications:&\\

 5.1.&Configuration of maintenance devices for satellite mission&\cite{fur86}\\

 5.2.&Balanced combinatorial cooperative games& \cite{faig93,faig98,wo95}\\

 5.3.&Technology mapping in field-programmable gate array&\cite{gar79}\\
     &semiconductor chip design &\\

 5.4.&Production scheduling &  \cite{bennell13,hop99}\\

\hline
\end{tabular}
\end{center}

\newpage

\begin{center}
{\bf Table 7.} Main algorithmic approaches, part I: basic methods  \\
\begin{tabular}{| c | l| l |}
\hline
 No.  & Solving approach  &Some source(s) \\

\hline
 I.& Fitting algorithms (i.e., classical ones) and their combinations:&\cite{coff97,cof07,cof13,gar79}\\
 1.1.& Next Fit (NF) algorithm&\cite{zhu11}\\
 1.2.& Next-fit (NFD)
  decreasing algorithm&\cite{zhu11}\\

 1.3.& First-Fit (FF) (on-line)&\cite{bak85,dosa14}\\

 1.4.& First-Fit decreasing (FFD) (off-line)&\cite{bak85,dosa13,sim94}\\

 1.5.& Best-Fit (BF) (on-line) &\cite{bennell13,dosa14,zhu11}\\

 1.6.& Best-Fit
 decreasing (BFD) algorithm (off-line)&\cite{sim94,zhu11}\\

 1.7.&Worst Fit (WF) algorithm (makespan context)&\cite{cof07}\\

 1.8.&Worst Fit decreasing (makespan context)&\cite{cof07}\\

 1.9.&Shelf algorithms
 (for 2D bin packing problems)
 &\cite{baker97,csirik97}\\

\hline
 II.&Exact enumerative methods:&\\

 2.1.& Surveys &\cite{del15}\\

 2.2.&Branch-and-bound algorithms&\cite{eil71,labbe95,muk00,per16,schol97,val99} \\



 2.3.&Branch-and-price algorithms&\cite{peet06,vance98,vance94}\\

 2.4.&Exact column generation and branch-and-bound method
  &\cite{val99,vance94}\\

 2.5.&Bin completion algorithm
 &\cite{fuk07}\\
     &(bin-oriented branch-and-bound strategy)&\\

\hline
 III.& Basic approximation algorithms:&\\

 3.1.&Surveys &\cite{coff97,cof13,del15,rood86}\\

 3.2.&Near-optimal algorithms for bin packing&\cite{john73,rood86}\\

 3.3.&Fast algorithms for bin packing &\cite{john74a}\\

 3.4.& Linear-time approximation algorithms for bin packing
 &\cite{zhang99} \\

 3.5.&Efficient approximation scheme&\cite{karm82}\\

 3.6.&Efficient approximation scheme for variable sized bin packing&\cite{murg87}\\

 3.7.& Asymptotic Polynomial Time Approximation Scheme (APTAS)
  &\cite{bal15,bein08,fern81,twigl15}\\

 3.8.& Asymptotic Fully Polynomial Time Approximation Scheme
  &\cite{bal15,karm82}\\

 & (AFPTAS) &\\

 3.9.& Augmented asymptotic PTAS &\cite{corr08}\\

 3.10.&Robust APTAS (for classical bin packing)&\cite{eps09}\\

 3.11.&Approximation schemes for multidimensional problems &\cite{bansal06,bansal09}\\

\hline
\end{tabular}
\end{center}

\begin{center}
\begin{picture}(116,71)
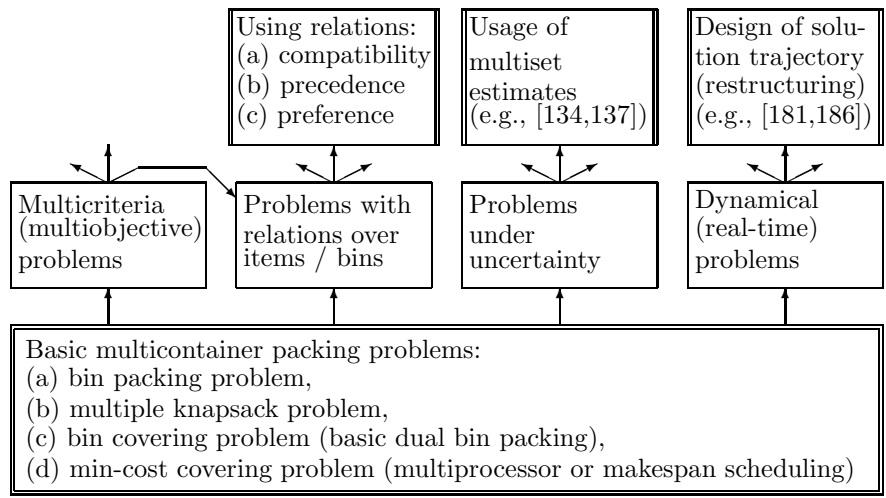
\put(10.5,00){\makebox(0,0)[bl]{Fig. 7. Examined extension trends
 in bin packing problems}}

\put(90,52){\line(1,0){26}} \put(90,70){\line(1,0){26}}
\put(90,52){\line(0,1){18}} \put(116,52){\line(0,1){18}}
\put(90.5,52){\line(0,1){18}} \put(115.5,52){\line(0,1){18}}

\put(91,66){\makebox(0,0)[bl]{Design of solu-}}
\put(91,62){\makebox(0,0)[bl]{tion trajectory}}
\put(91,58){\makebox(0,0)[bl]{(restructuring)}}
\put(91,54){\makebox(0,0)[bl]{(e.g., \cite{lev11restr,lev15restr})
}}

\put(90,33){\line(1,0){26}} \put(90,47){\line(1,0){26}}
\put(90,33){\line(0,1){14}} \put(116,33){\line(0,1){14}}

\put(91,43){\makebox(0,0)[bl]{Dynamical}}
\put(91,39){\makebox(0,0)[bl]{(real-time)}}
\put(91,35){\makebox(0,0)[bl]{problems}}
\put(60,52){\line(1,0){26}} \put(60,70){\line(1,0){26}}
\put(60,52){\line(0,1){18}} \put(86,52){\line(0,1){18}}
\put(60.5,52){\line(0,1){18}} \put(85.5,52){\line(0,1){18}}

\put(61,66){\makebox(0,0)[bl]{Usage of  }}
\put(61,62){\makebox(0,0)[bl]{multiset }}
\put(61,58){\makebox(0,0)[bl]{estimates }}
\put(61,54){\makebox(0,0)[bl]{(e.g., [134,137])}}

\put(60,33){\line(1,0){26}} \put(60,47){\line(1,0){26}}
\put(60,33){\line(0,1){14}} \put(86,33){\line(0,1){14}}

\put(61,43){\makebox(0,0)[bl]{Problems}}
\put(61,39){\makebox(0,0)[bl]{under }}
\put(61,35){\makebox(0,0)[bl]{uncertainty }}

\put(29,52){\line(1,0){28}} \put(29,70){\line(1,0){28}}
\put(29,52){\line(0,1){18}} \put(57,52){\line(0,1){18}}
\put(29.5,52){\line(0,1){18}} \put(56.5,52){\line(0,1){18}}

\put(30,66){\makebox(0,0)[bl]{Using relations:}}
\put(30,62){\makebox(0,0)[bl]{(a) compatibility }}
\put(30,58){\makebox(0,0)[bl]{(b) precedence}}
\put(30,54){\makebox(0,0)[bl]{(c) preference}}

\put(30,33){\line(1,0){26}} \put(30,47){\line(1,0){26}}
\put(30,33){\line(0,1){14}} \put(56,33){\line(0,1){14}}

\put(31,43){\makebox(0,0)[bl]{Problems with}}
\put(31,39){\makebox(0,0)[bl]{relations over }}
\put(31,35){\makebox(0,0)[bl]{items / bins}}
\put(00,33){\line(1,0){26}} \put(00,47){\line(1,0){26}}
\put(00,33){\line(0,1){14}} \put(26,33){\line(0,1){14}}

\put(01,43){\makebox(0,0)[bl]{Multicriteria}}
\put(01,39){\makebox(0,0)[bl]{(multiobjective) }}
\put(01,35){\makebox(0,0)[bl]{problems }}
\put(13,47){\vector(0,1){5}}

\put(13,47){\vector(-2,1){5}}

\put(13,47){\line(2,1){4}} \put(17,49){\line(1,0){09}}
\put(26,49){\vector(1,-1){4}}


\put(43,47){\vector(0,1){5}}

\put(43,47){\vector(-2,1){5}} \put(43,47){\vector(2,1){5}}

\put(73,47){\vector(0,1){5}}

\put(73,47){\vector(-2,1){5}} \put(73,47){\vector(2,1){5}}

\put(103,47){\vector(0,1){5}}

\put(103,47){\vector(-2,1){5}} \put(103,47){\vector(2,1){5}}


\put(13,28){\vector(0,1){5}} \put(43,28){\vector(0,1){5}}
\put(73,28){\vector(0,1){5}} \put(103,28){\vector(0,1){5}}

\put(00,05.5){\line(1,0){116}} \put(00,28){\line(1,0){116}}
\put(00,05.5){\line(0,1){22.5}} \put(116,05.5){\line(0,1){22.5}}

\put(0.5,06){\line(1,0){115}} \put(0.5,27.5){\line(1,0){115}}
\put(0.5,06){\line(0,1){21.5}} \put(115.5,06){\line(0,1){21.5}}

\put(02,23){\makebox(0,0)[bl]{Basic multicontainer packing
  problems:}}

\put(02,19){\makebox(0,0)[bl]{(a) bin packing problem,}}

\put(02,15){\makebox(0,0)[bl]{(b) multiple knapsack problem,}}

\put(02,11){\makebox(0,0)[bl]{(c) bin covering problem (basic dual
bin packing),}}

\put(02,07){\makebox(0,0)[bl]{(d) min-cost covering problem
(multiprocessor or makespan scheduling)}}

\end{picture}
\end{center}
%

\newpage
\begin{center}
 {\bf Table 8.} Main algorithmic approaches, part II:  heuristics  \\
\begin{tabular}{| c | l| l |}
 \hline
 No.  & Solving approach  &Some source(s) \\

 \hline

 \hline
 IV.& Heuristics: &\\

 4.1.&Surveys and heuristics comparison  & \cite{cof02a,dow91,haou09}\\

 4.2.&Basic heuristics  &\cite{dai94,gend04,gent98,hop01,lodi99b,lopez13} \\

 4.3.&Local search algorithms &\cite{levine04,oso03}\\

 4.4.&Greedy procedures  &\cite{chwa11}\\

 4.5.&Variable neighborhood search procedures&\cite{chwa11,flez02}\\

 4.6.&Dynamic programming based heuristics&\cite{per16} \\

 4.7.& Simulated annealing based algorithms&\cite{beis05,kamp88,szy94}\\

 4.8.&Tabu search algorithms & \cite{alvim04,lodi04,schol97}\\

 4.9.& GRASP algorithms&\cite{layeb12}\\

 4.10.&Ant colony algorithms&\cite{levine04}\\

 4.11.&Quantum inspired cuckoo search algorithms&\cite{layeb12a}\\

 4.12.&Set-covering-based heuristics&\cite{bansal09,mon06}\\

 4.13.& Average-weight-controlled bin-oriented heuristics
  & \cite{flez11}\\

 4.14.& Bottom-left bin packing heuristic (for 2D problem)
 &\cite{chaz83}\\

 4.15.&Heuristic for 2D and 3D large bin packing
 &\cite{mack10}\\

\hline
 V.&Hybrid approaches, metaheuristics and hyper-heuristics: &\\

 5.1.&Hybrid approach, metaheuristrics for 2D bin packing&\cite{cui16,hong14,hop01,hop01a,lodi99b}\\

 5.2.&Hyper-heuristics, generalized hyper-heuristics&\cite{sim12,tera10}\\

 5.3.&Unified hyper-heuristic framework&\cite{lopez14}\\

 5.4.&Combinations of  evolutionary algorithms and hyper-heuristics
  &\cite{burke06,lopez11,ross03}\\

 5.5.&
 Combination of Lagrangian relaxation and column generation
  & \cite{elh15}\\

\hline
\end{tabular}
\end{center}

\begin{center}
{\bf Table 9.} Main algorithmic approaches, part III: online and evolutionary methods  \\
\begin{tabular}{| c | l| l |}
\hline
 No.  & Solving approach  &Some source(s) \\

\hline

 VI.&Online and dynamic algorithms for bin packing:&\\

 6.1.&Survey of online algorithms for bin packing&\cite{eps12}\\

 6.2.&Fully dynamic algorithms for bin packing
  &\cite{ivk98}\\

 6.3.&Simple on-line bin-packing algorithm&\cite{lee85}\\

 6.4.&Onlne algorithms for variable sized bin packing&\cite{csirik89,seid01,woeg99} \\

 6.5.&On-line algorithms for dual version of bin packing
 &\cite{csirik88}\\

 6.6.&On-line algorithm for multidimensional bin packing
  &\cite{csirik93}\\

\hline
 VII.&Evolutionary approaches:&\\

 7.1.&Genetic algorithms/evolutionary based heuristics
 &\cite{bennell13,bhat04,burke06,hop99,krog95,sta08}\\

 7.2.&Genetic algorithms  in 2D  packing problems&\cite{jain98}\\

 7.2.&Mixed simulated annealing-genetic algorithm
   &\cite{leung03}\\

  &for 2D orthogonal packing&\\

 7.3.&Evolutionary particle swam optimization
 &\cite{liud08}\\

  &for multiobjective bin packing&\\

 7.4.& Hybrid genetic algorithms &\cite{reev96}\\

 7.5.&Grouping genetic algorithms&\cite{ulker08}\\

 7.6.&Grouping genetical algorithm with controlled gene transmission&\cite{qui15}\\

 7.7.& Hybrid grouping genetic algorithms&\cite{falk96}\\

 7.8.&Nature inspired genetic algorithms&\cite{roh10}\\

 7.9.&Histogram-matching approach to the evolution of &\cite{poli07}\\

 & bin-packing strategies
  for discrete sizes of item/bins&\\

 7.10.&Combinations of  evolutionary algorithms and hyper-heuristics
  &\cite{burke06,lopez11,ross03}\\

\hline
\end{tabular}
\end{center}

\newpage

\section{Preliminary information}

\subsection{Basic problem formulations}

 The classical formal statement of BPP is the following
 (e.g., \cite{john73,john74a,john74,ull71}).
 Given a bin \(S\) of size \(V\)
 and a list of \(n\) items with sizes \(a_{1},...,a_{n}\)
 to pack.

~~

 Find an integer number of bins \(B\) and a \(B\)-partition
 \(S_{1} \bigcup  ... \bigcup S_{B}\)
 of set \(\{1,...,n\}\) such that
 \(\sum_{i\in S_{k}} a_{i} \leq V\) for all \(k=1,...,B.\)

~~

 A solution is optimal if it has minimal \(B\).
 The \(B\)-value for an optimal solution is denoted OPT below.

 A possible integer linear formulation of the problem is \cite{mar90}:
 \[\min B = \sum_{i=1}^{n} y_{i}\]
 \[s.t.~~~~~ B\geq 1, ~~~~
%
 \sum_{j=1}^{n} a_{j} x_{ij}  \leq  V y_{i},  ~ \forall i \in \{1,...,n\}
%
 ~~~~ \sum_{i=1}^{n} x_{ij}  =  1,  ~ \forall j \in \{1,...,n\}\]
 \[ y_{i} \in \{0,1\}, ~  \forall i \in \{1,...,n\} ~~~~
%
  x_{ij} \in \{0,1\}, ~  \forall i \in \{1,...,n\},  ~ \forall j \in \{1,...,n\}\]
 where \(y_{i} = 1\) if bin \(i\) is used and \(x_{ij} = 1\) if
 item \(j\) is put into bin \(i\).

\subsection{Maximizing the number of packed items (inverse problems)}

 The inverse bin packing problem is targeted to maximization
 of the number of packed items.
 Here, two basic kinds of the problems have been considered:

 (i) maximization of the number of packed items
 (the number of bins is fixed)
 (e.g., \cite{cof78});

 (ii) ``maximization'' of the total preference estimate for packed items
 (the number of bins is fixed,
 (preference relation over item set)
 (e.g., \cite{fur86}).

 The description of inverse bin packing problem
 will be examined in further section.

\subsection{Interval multiset estimates}

 Interval multiset estimates have been suggested by M.Sh. Levin
 in \cite{lev12a}.
 A brief description of interval multiset estimates is the
 following
 \cite{lev12a,lev15}.
 The approach consists in assignment of elements (\(1,2,3,...\))
 into an ordinal scale \([1,2,...,l]\).
 As a result, a multi-set based estimate is obtained,
 where a basis set involves all levels of the ordinal scale:
 \(\Omega = \{ 1,2,...,l\}\) (the levels are linear ordered:
 \(1 \succ 2 \succ 3 \succ ...\)) and
 the assessment problem (for each alternative)
 consists in selection of a multiset over set \(\Omega\) while taking into
 account two conditions:

 {\it 1.} cardinality of the selected multiset equals a specified
 number of elements \( \eta = 1,2,3,...\)
 (i.e., multisets of cardinality \(\eta \) are considered);

 {\it 2.} ``configuration'' of the multiset is the following:
 the selected elements of \(\Omega\) cover an interval over scale \([1,l]\)
 (i.e., ``interval multiset estimate'').

 Thus, an estimate \(e\) for an alternative \(A\) is
 (scale \([1,l]\), position-based form or position form):
 \(e(A) = (\eta_{1},...,\eta_{\iota},...,\eta_{l})\),
 where \(\eta_{\iota}\) corresponds to the number of elements at the
 level \(\iota\) (\(\iota = \overline{1,l}\)), or
 \(e(A) = \{ \overbrace{1,...,1}^{\eta_{1}},\overbrace{2,...2}^{\eta_{2}},
 \overbrace{3,...,3}^{\eta_{3}},...,\overbrace{l,...,l}^{\eta_{l}}
 \}\).
 The number of multisets of cardinality \(\eta\),
 with elements taken from a finite set of cardinality \(l\),
 is called the
 ``multiset coefficient'' or ``multiset number''
  (\cite{knuth98,yager86}):
 ~~\( \mu^{l,\eta} =
   \frac{l(l+1)(l+2)... (l+\eta-1) } {\eta!}
   \).
 This number corresponds to possible estimates
 (without taking into account interval condition 2).
 In the case of condition 2,
 the number of estimates is decreased.
 Generally, assessment problems based on interval multiset estimates
 can be denoted as follows: ~\(P^{l,\eta}\).

 A poset-like scale of interval multiset estimates for assessment problem \(P^{3,3}\)
 is presented
 in Fig. 8.

\begin{center}
\begin{picture}(84,102)

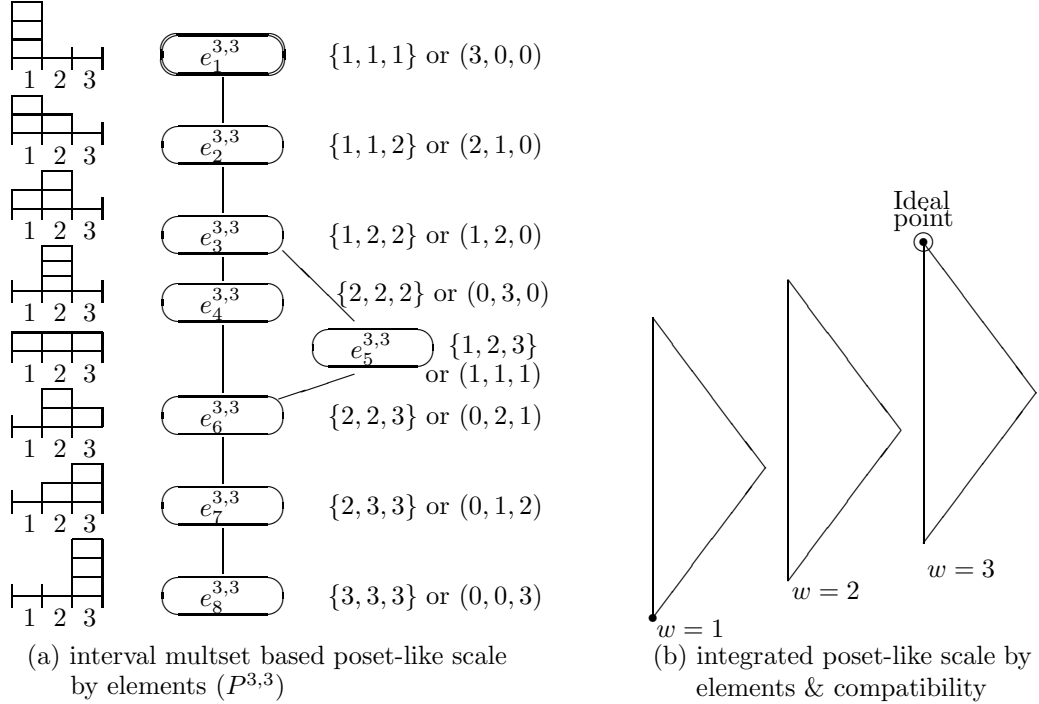
\put(22,00){\makebox(0,0)[bl]{Fig. 8. Multiset based
 scale, estimates (based on \cite{lev12a,lev15}) }}

\put(02,09){\makebox(0,0)[bl]{(a) interval multset based
poset-like scale}}

\put(07.8,05){\makebox(0,0)[bl]{by elements (\(P^{3,3}\))}}

\put(25,88.7){\makebox(0,0)[bl]{\(e^{3,3}_{1}\) }}

\put(28,91){\oval(16,5)} \put(28,91){\oval(16.5,5.5)}


\put(42,89){\makebox(0,0)[bl]{\(\{1,1,1\}\) or \((3,0,0)\) }}

\put(00,90.5){\line(0,1){07.5}} \put(04,90.5){\line(0,1){07.5}}

\put(00,93){\line(1,0){04}} \put(00,95.5){\line(1,0){04}}
\put(00,98){\line(1,0){4}}

\put(00,90.5){\line(1,0){12}}

\put(00,89){\line(0,1){3}} \put(04,89){\line(0,1){3}}
\put(08,89){\line(0,1){3}} \put(12,89){\line(0,1){3}}

\put(01.5,86.5){\makebox(0,0)[bl]{\(1\)}}
\put(05.5,86.5){\makebox(0,0)[bl]{\(2\)}}
\put(09.5,86.5){\makebox(0,0)[bl]{\(3\)}}


\put(28,82){\line(0,1){6}}

\put(25,76.7){\makebox(0,0)[bl]{\(e^{3,3}_{2}\) }}

\put(28,79){\oval(16,5)}


\put(42,77){\makebox(0,0)[bl]{\(\{1,1,2\}\) or \((2,1,0)\) }}

\put(00,80.5){\line(0,1){05}} \put(04,80.5){\line(0,1){05}}
\put(8,80.5){\line(0,1){02.5}}

\put(00,83){\line(1,0){8}} \put(00,85.5){\line(1,0){4}}

\put(00,80.5){\line(1,0){12}}

\put(00,79){\line(0,1){3}} \put(04,79){\line(0,1){3}}
\put(08,79){\line(0,1){3}} \put(12,79){\line(0,1){3}}

\put(01.5,76.5){\makebox(0,0)[bl]{\(1\)}}
\put(05.5,76.5){\makebox(0,0)[bl]{\(2\)}}
\put(09.5,76.5){\makebox(0,0)[bl]{\(3\)}}


\put(28,70){\line(0,1){6}}

\put(25,64.7){\makebox(0,0)[bl]{\(e^{3,3}_{3}\) }}

\put(28,67){\oval(16,5)}


\put(42,65){\makebox(0,0)[bl]{\(\{1,2,2\}\) or \((1,2,0)\) }}

\put(00,70.5){\line(0,1){02.5}} \put(04,70.5){\line(0,1){05}}
\put(8,70.5){\line(0,1){05}}

\put(00,73){\line(1,0){8}} \put(04,75.5){\line(1,0){4}}

\put(00,70.5){\line(1,0){12}}

\put(00,69){\line(0,1){3}} \put(04,69){\line(0,1){3}}
\put(08,69){\line(0,1){3}} \put(12,69){\line(0,1){3}}

\put(01.5,66.5){\makebox(0,0)[bl]{\(1\)}}
\put(05.5,66.5){\makebox(0,0)[bl]{\(2\)}}
\put(09.5,66.5){\makebox(0,0)[bl]{\(3\)}}


\put(28,61){\line(0,1){3}}

\put(25,55.7){\makebox(0,0)[bl]{\(e^{3,3}_{4}\) }}

\put(28,58){\oval(16,5)}


\put(43,57.4){\makebox(0,0)[bl]{\(\{2,2,2\}\) or \((0,3,0)\) }}

\put(04,59.5){\line(0,1){06}} \put(08,59.5){\line(0,1){06}}

\put(04,61.5){\line(1,0){04}} \put(04,63.5){\line(1,0){04}}
\put(04,65.5){\line(1,0){04}}

\put(00,59.5){\line(1,0){12}}

\put(00,58){\line(0,1){3}} \put(04,58){\line(0,1){3}}
\put(08,58){\line(0,1){3}} \put(12,58){\line(0,1){3}}

\put(01.5,55.5){\makebox(0,0)[bl]{\(1\)}}
\put(05.5,55.5){\makebox(0,0)[bl]{\(2\)}}
\put(09.5,55.5){\makebox(0,0)[bl]{\(3\)}}


\put(28,46){\line(0,1){9}}

\put(25,40.7){\makebox(0,0)[bl]{\(e^{3,3}_{6}\) }}

\put(28,43){\oval(16,5)}


\put(42,41){\makebox(0,0)[bl]{\(\{2,2,3\}\) or \((0,2,1)\) }}

\put(04,41.5){\line(0,1){05}} \put(08,41.5){\line(0,1){05}}
\put(12,41.5){\line(0,1){02.5}}

\put(04,44){\line(1,0){8}} \put(04,46.5){\line(1,0){4}}

\put(00,41.5){\line(1,0){12}}

\put(00,40){\line(0,1){3}} \put(04,40){\line(0,1){3}}
\put(08,40){\line(0,1){3}} \put(12,30){\line(0,1){3}}

\put(01.5,37.5){\makebox(0,0)[bl]{\(1\)}}
\put(05.5,37.5){\makebox(0,0)[bl]{\(2\)}}
\put(09.5,37.5){\makebox(0,0)[bl]{\(3\)}}


\put(28,34){\line(0,1){6}}

\put(25,28.7){\makebox(0,0)[bl]{\(e^{3,3}_{7}\) }}

\put(28,31){\oval(16,5)}


\put(42,29){\makebox(0,0)[bl]{\(\{2,3,3\}\) or \((0,1,2)\) }}

\put(04,31.5){\line(0,1){02.5}} \put(08,31.5){\line(0,1){05}}
\put(12,31.5){\line(0,1){05}}

\put(04,34){\line(1,0){8}} \put(08,36.5){\line(1,0){4}}

\put(00,31.5){\line(1,0){12}}

\put(00,30){\line(0,1){3}} \put(04,30){\line(0,1){3}}
\put(08,30){\line(0,1){3}} \put(12,30){\line(0,1){3}}

\put(01.5,27.5){\makebox(0,0)[bl]{\(1\)}}
\put(05.5,27.5){\makebox(0,0)[bl]{\(2\)}}
\put(09.5,27.5){\makebox(0,0)[bl]{\(3\)}}


\put(28,22){\line(0,1){6}}

\put(25,16.7){\makebox(0,0)[bl]{\(e^{3,3}_{8}\) }}

\put(28,19){\oval(16,5)}


\put(42,17){\makebox(0,0)[bl]{\(\{3,3,3\}\) or \((0,0,3)\) }}

\put(08,19){\line(0,1){07.5}} \put(12,19){\line(0,1){07.5}}
\put(08,21.5){\line(1,0){4}} \put(08,24){\line(1,0){4}}
\put(08,26.5){\line(1,0){4}}

\put(00,19){\line(1,0){12}}

\put(00,017.5){\line(0,1){3}} \put(04,017.5){\line(0,1){3}}
\put(08,017.5){\line(0,1){3}} \put(12,017.5){\line(0,1){3}}

\put(01.5,15){\makebox(0,0)[bl]{\(1\)}}
\put(05.5,15){\makebox(0,0)[bl]{\(2\)}}
\put(09.5,15){\makebox(0,0)[bl]{\(3\)}}


\put(45.5,55.5){\line(-1,1){09.5}}

\put(45.5,48.5){\line(-3,-1){10}}

\put(45,49.7){\makebox(0,0)[bl]{\(e^{3,3}_{5}\) }}

\put(48,52){\oval(16,5)}


\put(58,50.5){\makebox(0,0)[bl]{\(\{1,2,3\}\) }}
\put(55,46.5){\makebox(0,0)[bl]{or \((1,1,1)\) }}

\put(00,51.5){\line(0,1){02.5}} \put(04,51.5){\line(0,1){02.5}}
\put(8,51.5){\line(0,1){02.5}} \put(12,51.5){\line(0,1){02.5}}

\put(00,54){\line(1,0){12}}

\put(00,51.5){\line(1,0){12}}

\put(00,50){\line(0,1){3}} \put(04,50){\line(0,1){3}}
\put(08,50){\line(0,1){3}} \put(12,50){\line(0,1){3}}

\put(01.5,47.5){\makebox(0,0)[bl]{\(1\)}}
\put(05.5,47.5){\makebox(0,0)[bl]{\(2\)}}
\put(09.5,47.5){\makebox(0,0)[bl]{\(3\)}}


\end{picture}
%
\begin{picture}(56,59)

\put(00,09){\makebox(0,0)[bl]{(b) integrated poset-like scale by}}

\put(05.8,05){\makebox(0,0)[bl]{elements \& compatibility}}

\put(00,16){\circle*{0.9}}

\put(00,16){\line(0,1){40}} \put(00,16){\line(3,4){15}}
\put(00,56){\line(3,-4){15}}

\put(18,21){\line(0,1){40}} \put(18,21){\line(3,4){15}}
\put(18,61){\line(3,-4){15}}

\put(36,26){\line(0,1){40}} \put(36,26){\line(3,4){15}}
\put(36,66){\line(3,-4){15}}


\put(36,66){\circle*{1}} \put(36,66){\circle{2.5}}

\put(32,70.5){\makebox(0,0)[bl]{Ideal}}
\put(32,67.5){\makebox(0,0)[bl]{point}}

\put(00.5,13.5){\makebox(0,0)[bl]{\(w=1\)}}
\put(18.5,18.5){\makebox(0,0)[bl]{\(w=2\)}}
\put(36.5,21.5){\makebox(0,0)[bl]{\(w=3\)}}

\end{picture}
\end{center}

 Fig. 8a illustrates the scale-poset and estimates for problem
 \(P_{3,3}\)
 (assessment over scale \([1,3]\)
 with three elements, estimates \((2,0,2\)
 and \((1,0,2)\) are not used)
 \cite{lev12a,lev15}.
 For evaluation of multi-component system,
 multi-component poset-like scale
 composed from several poset-like scale  may be used
 \cite{lev12a,lev15}.
 Fig. 8b depicts
 the integrated poset-like scale for tree-component system
 (ordinal scale for system component compatibility is \([0,1,2,3]\)).

 The following operations over multiset estimates
 are used \cite{lev12a,lev15} as well:
 integration, vector-like proximity, aggregation, and alignment.


 Integration of estimates (mainly, for composite systems)
 is based on summarization of the estimates by components (i.e.,
 positions).
 Let us consider \(n\) estimates (position form):~~
 estimate \(e^{1} = (\eta^{1}_{1},...,\eta^{1}_{\iota},...,\eta^{1}_{l})
 \),
  {\bf . . .},
 estimate \(e^{\kappa} = (\eta^{\kappa}_{1},...,\eta^{\kappa}_{\iota},...,\eta^{\kappa}_{l})
 \),
  {\bf . . .},
 estimate \(e^{n} = (\eta^{n}_{1},...,\eta^{n}_{\iota},...,\eta^{n}_{l})
 \).
 Then, the integrated estimate is:~
 estimate \(e^{I} = (\eta^{I}_{1},...,\eta^{I}_{\iota},...,\eta^{I}_{l})
 \),
 where
 \(\eta^{I}_{\iota} = \sum_{\kappa=1}^{n} \eta^{\kappa}_{\iota} ~~ \forall
 \iota = \overline{1,l}\).
 In fact, the operation \(\biguplus\) is used for multiset estimates:
 \(e^{I} = e^{1} \biguplus ... \biguplus e^{\kappa} \biguplus ... \biguplus e^{n}\).


 Further, vector-like proximity is described.
  Let \(A_{1}\) and \(A_{2}\) be two alternatives
 with corresponding
 interval multiset estimates
 \(e(A_{1})\), \(e(A_{2})\).
  Vector-like proximity for the alternatives above is:
 ~~\(\delta ( e(A_{1}), e(A_{2})) = (\delta^{-}(A_{1},A_{2}),\delta^{+}(A_{1},A_{2}))\),
 where vector components are:
 (i) \(\delta^{-}\) is the number of one-step changes:
 element of quality \(\iota + 1\) into element of quality \(\iota\) (\(\iota = \overline{1,l-1}\))
 (this corresponds to ``improvement'');
 (ii) \(\delta^{+}\) is the number of one-step changes:
 element of quality \(\iota\) into element of quality  \(\iota+1\) (\(\iota = \overline{1,l-1}\))
 (this corresponds to ``degradation'').
 It is assumed:
 ~\( | \delta ( e(A_{1}), e(A_{2})) | = | \delta^{-}(A_{1},A_{2}) | + |\delta^{+}(A_{1},A_{2})|
 \).
%


 Now let us consider median estimates (aggregation)
  for the
 specified set of initial estimates
 (traditional approach).
 Let \(E = \{ e_{1},...,e_{\kappa},...,e_{n}\}\)
 be the set of specified estimates
 (or a corresponding set of specified alternatives),
 let \(\overline{D} \)
 be the set of all possible estimates
 (or a corresponding set of possible alternatives)
 (\( E  \subseteq \overline{D} \)).
%
  Thus, the median estimates
  (``generalized median'' \(M^{g}\) and ``set median'' \(M^{s}\)) are:
 ~~\(M^{g} =   \arg \min_{M \in \overline{D}}~
   \sum_{\kappa=1}^{n} ~  | \delta (M, e_{\kappa}) |; ~~
 M^{s} =   \arg \min_{M\in E} ~
   \sum_{\kappa=1}^{n} ~ | \delta (M, e_{\kappa}) |\).

 In recent decade,
 the significance of
  multiset studies and applications has been increased.
 Some recent studies in multisets and their
 applications are pointed out in Table 9.

\newpage
\begin{center}
 {\bf Table 9.} Studies in multisets and their applications\\
\begin{tabular}{| c | l| l | l |}
\hline
 No.  &Research direction(s)  &Source(s) \\
\hline
 I.&Formal models, definitions:&\\

 1.1.&Basic definitions, development of multiset theory&  \cite{bliz88,bliz91,knuth98,singh08,yager86}\\

 1.2.&Mathematics of multisets (axiomatic view, operations between& \cite{bey09,dovier98,syr00}\\
   & multisets)&\\

 1.3.&Multiset automata (Chomsky-like hierarchy of multiset grammars&\cite{csu00} \\
    &in terms of mutliset automata)&\\

 1.4.& High-level framework for the definition of visual languages
 & \cite{marr94}\\
  & (constraint multiset grammars)&\\

 1.5.&Fuzzy multisets, their generalization, soft multisets theory&\cite{alk11,bed12,miy01,miy05}\\


 1.6.&Tolerance multisets &\cite{marc01}\\

 1.7.&Multiset metric spaces& \cite{ibr12,pet08}\\

 1.8.&Framework for multiset merging& \cite{bron12}\\

 1.9.&Interval multiset estimates, operations over multisets&\cite{lev12a,lev15}\\
     &(e.g., proximity, summarization, aggregation)&\\

 1.10.& Perturbation of multisets
  (measure of remoteness between multisets)&\cite{kraw15}\\

 1.11.&Multiset processing (general)&\cite{calude01}\\

\hline
 II.&Some applications:&\\

 2.1.&Multisets in database systems&\cite{lamp01}\\

 2.2.&Neural network processing of multiset data &\cite{mcg07}\\

 2.3.&Programs as multiset transformations&\cite{bana88,bana93}\\

 2.4.&Multiset rewriting systems &\cite{alh11,dur04}\\

 2.5.&Proving termination with multiset ordering&\cite{der79,hof92}\\

 2.6.&Automatic construction of user interfaces&\cite{sen95}\\

 2.7.&Clustering&\cite{miy03,pet97,pet08}\\

 2.8.&Classification (e.g., classification of credit cardholders)
  &\cite{pet06}\\

 2.9.&Applications in decision making
 (e.g., multicriteria ranking/sorting)
   &\cite{bed12,pet97,pet08}\\

 2.10.&Processing of data streams&\cite{babk03,guha00,guha01}\\

 2.11.&Evaluation of composite system(s)/alternative(s)&\cite{lev98,lev06,lev12a,lev15}\\

 2.12.&Knapsack problem                &\cite{lev12a,lev15}\\

 2.13.&Multiple choice knapsack problem&\cite{lev12a,lev15}\\

 2.14.&Combinatorial synthesis (morphological system design) &\cite{lev98,lev06,lev12a,lev15}\\

\hline
\end{tabular}
\end{center}

\subsection{Support model: morphological design with ordinal and interval multiset estimates}

 A brief description  of combinatorial synthesis
 (Hierarchical Morphological Multicriteria Design - HMMD)
 with ordinal estimates of design alternatives
 is the following
 (\cite{lev98,lev06,lev12a,lev15}).
 An examined composite
 (modular, decomposable) system consists
 of components and their interconnection or compatibility (IC).
 Basic assumptions of HMMD are the following:
 ~{\it (a)} a tree-like structure of the system;
 ~{\it (b)} a composite estimate for system quality
     that integrates components (subsystems, parts) qualities and
    qualities of IC (compatibility) across subsystems;
 ~{\it (c)} monotonic criteria for the system and its components;
 ~{\it (d)} quality of system components and IC are evaluated on the basis
    of coordinated ordinal scales.
 The designations are:
  ~(1) design alternatives (DAs) for leaf nodes of the model;
  ~(2) priorities of DAs (\(\iota = \overline{1,l}\);
      \(1\) corresponds to the best one);
  ~(3) ordinal compatibility for each pair of DAs
  (\(w=\overline{1,\nu}\); \(\nu\) corresponds to the best one).
 Let \(S\) be a system consisting of \(m\) parts (components):
 \(R(1),...,R(i),...,R(m)\).
 A set of design alternatives
 is generated for each system part above.
 The problem is:

~~

 {\it Find a composite design alternative}
 ~~ \(S=S(1)\star ...\star S(i)\star ...\star S(m)\)~~
 {\it of DAs (one representative design alternative}
 ~\(S(i)\)
 {\it for each system component/part}
  ~\(R(i)\), \(i=\overline{1,m}\)
  {\it )}
 {\it with non-zero compatibility}
 {\it between design alternatives.}

~~

 A discrete ``space'' of the system excellence
 (a poset)
 on the basis of the following vector is used:
 ~~\(N(S)=(w(S);e(S))\),
 ~where \(w(S)\) is the minimum of pairwise compatibility
 between DAs which correspond to different system components
 (i.e.,
 \(~\forall ~R_{j_{1}}\) and \( R_{j_{2}}\),
 \(1 \leq j_{1} \neq j_{2} \leq m\))
 in \(S\),
 ~\(e(S)=(\eta_{1},...,\eta_{\iota},...,\eta_{l})\),
 ~where \(\eta_{\iota}\) is the number of DAs of the \(\iota\)th quality in \(S\).
 Further,
  the problem is described as follows:
 \[ \max~ e(S), ~~~ \max~ w(S),
  ~~~~~~~ s.t.
  ~~ w(S) \geq 1  .\]
 As a result,
 we search for composite solutions
 which are nondominated by \(N(S)\)
 (i.e., Pareto-efficient).
 ``Maximization''  of \(e(S)\) is based on the corresponding poset.
 The considered combinatorial problem is NP-hard
 and an enumerative solving scheme is used.
%

%
 Here, combinatorial synthesis is based on usage of multiset
 estimates of design alternatives for system parts.
 For the resultant system \(S = S(1) \star ... \star S(i) \star ... \star S(m) \)
 the same type of the multiset estimate is examined:
  an aggregated estimate (``generalized median'')
  of corresponding multiset estimates of its components
 (i.e., selected DAs).
 Thus, \( N(S) = (w(S);e(S))\), where
 \(e(S)\) is the ``generalized median'' of estimates of the solution
 components.
 Finally, the modified problem is:
 \[ \max~ e(S) = M^{g} =
  \arg \min_{M \in \overline{D} }~~
  \sum_{i=1}^{m} ~ |\delta (M, e(S_{i})) |,
 ~~~ \max~ w(S),
 ~~~~~~~ s.t.
  ~~ w(S) \geq 1  .\]
 Enumeration methods or heuristics can be  used
 (\cite{lev98,lev06,lev12a,lev15}).

\newpage
\section{Problems with multiset estimates}

\subsection{Some combinatorial optimization problems with
 multiset estimates}

\subsubsection{Knapsack problem with multiset estimates}

 The basic knapsack problem (i.e., ``\(0-1\)  knapsack problem'')
 is
 (e.g., \cite{gar79,keller04,mar90}):
 (i) given item set \( A= \{ 1,...,i,...,m \}\)
 with parameters   \( \forall i \in A \):
  profit (or utility) \(\gamma_{i}\),
 resource requirement (e.g., weight) \(a_{i}\);
 (ii) given a resource (capacity) of knapsack \( b\).
 Thus, the model is as follows:
 \[\max\sum_{i=1}^{m} \gamma_{i} x_{i}
%
 ~~~~~~ s.t. ~~ \sum_{i=1}^{m} a_{i} x_{i} \leq b,
 ~~ x_{i} \in\{0,1\}, ~ i=\overline{1,m}\]
%
%
%
 where \(x_{i}=1\) if item \(i\) is selected,
  and \(x_{i}=0\) otherwise.
%
 Often nonnegative coefficients are assumed.

 In the case of multiset estimates of
 item ``utility'' \(e_{i}, ~i \in \{1,...,i,...,n\}\)
 (instead of \(\gamma_{i}\)),
 the following aggregated multiset estimate can be used
 for the objective function  (``maximization'')
  (e.g., \cite{lev12a,lev15}):
 (a) an aggregated multiset estimate as the ``generalized median'',
 (b) an aggregated multiset estimate as the ``set median'',
 and
 (c) an integrated  multiset estimate.
%
%
 Knapsack problem with multiset estimates
 and the integrated estimate for the solution is
 (solution \(S = \{ i | x_{i} = 1\}\)):
%
 \[ \max~ e(S) = \biguplus_{ i \in S=\{ i | x_{i}=1\}} ~  e_{i},
%
%
 ~~~~~~  s.t. ~~ \sum_{i=1}^{m}    a_{i} x_{i} \leq b;
%
%
 ~~ x_{i} \in \{0, 1\}.\]
 In the case of objective function based on median estimate
 for solution,
 the problem is:
%
%
 \[ \max~ e(S) =  \max~ M = arg \min_{M\in D} ~~
 | \biguplus_{ i \in S=\{ i | x_{i}=1\}} ~   \delta (M,e_{i}) |
%
%
%
 ~~~~~~  s.t. ~~ \sum_{i=1}^{m}    a_{i} x_{i} \leq b,
%
%
 ~~ x_{i} \in \{0, 1\}.\]

 In addition, it is reasonable to consider a new problem
 formulation while taking into account the number of the selected
 items (i.e. a special two-objective knapsack problem with multiset estimates)
 (solution \(S = \{ i | x_{i} = 1\}\)):
%
%
%
 \[ \max~ e(S) =  \max~ M = arg \min_{M\in D} ~~
 | \biguplus_{ i \in S=\{ i | x_{i}=1\}} ~   \delta (M,e_{i}) |
%
%
 ~~~~~~~ \max~ \sum_{i=1}^{n} x_{i}   \]
 \[  s.t. ~~ \sum_{i=1}^{m}    a_{i} x_{i} \leq b,
%
%
 ~~ x_{i} \in \{0, 1\}.\]
 Fig. 9 depicts the corresponding ``two''-dimensional
 space of solution quality.

\begin{center}
\begin{picture}(60,53)
\put(01,00){\makebox(0,0)[bl]{Fig. 9. ``2D''
 space of solution quality}}

\put(36,48){\makebox(0,0)[bl]{Total }}
\put(36,45){\makebox(0,0)[bl]{ideal}}
\put(36,42){\makebox(0,0)[bl]{point}}

\put(40,40){\circle*{1.5}} \put(40,40){\circle{2.5}}

\put(00,40){\line(1,0){40}}

\put(00,10){\vector(1,0){47}}

\put(48,12){\makebox(0,0)[bl]{Number of }}
\put(48,09){\makebox(0,0)[bl]{selected }}
\put(48,06){\makebox(0,0)[bl]{elements}}


\put(00,40){\circle*{1.5}}

\put(0,10){\line(0,1){30}} \put(0,10){\line(1,2){7.5}}
\put(0,40){\line(1,-2){7.5}}

\put(40,10){\line(0,1){30}} \put(40,10){\line(1,2){07.5}}
\put(40,40){\line(1,-2){07.5}}

\put(013,32){\makebox(0,0)[bl]{Lattice}}
\put(013,28){\makebox(0,0)[bl]{for}}
\put(013,25){\makebox(0,0)[bl]{element}}
\put(013,22){\makebox(0,0)[bl]{quality}}

\put(013,19){\makebox(0,0)[bl]{(poset-like}}
\put(013,16){\makebox(0,0)[bl]{scale)}}

\put(012.5,29){\line(-2,-1){09}}

\put(00,47){\makebox(0,0)[bl]{The ideal point}}
\put(00,44){\makebox(0,0)[bl]{by elements}}
\put(00,41){\makebox(0,0)[bl]{(e.g., ``median'')}}

\end{picture}
\end{center}

\subsubsection{Multiple choice problem with interval multiset estimates}

 In multiple choice problem,
 items are divided into groups (without  intersection)
 and items are selected in each group under total resource constraint
  (e.g., \cite{gar79,keller04,mar90}).
 Here, one item is selected in each group.
 This version of multiple choice problem is
 (Boolean variable \(x_{i,j}\) equals \(1\)
 if item \((i,j)\) is selected):
%
%
 \[\max\sum_{i=1}^{m} \sum_{j=1}^{q_{i}} \gamma_{ij} x_{ij}
%
 ~~~~~~~~ s.t. ~~ \sum_{i=1}^{m} \sum_{j=1}^{q_{i}} a_{ij} x_{ij} \leq b,
  ~~ \sum_{j=1}^{q_{i}} x_{ij} = 1,~ i=\overline{1,m},
  ~~~ x_{ij} \in \{0, 1\}.\]
%

 A special case of multiple choice problem is considered
 \cite{lev12a,lev15}:
 (1) multiset estimates of item ``utility''
 \(e_{ij}\) ~
  (\(i=\overline{1,m}\), ~
 \(j = \overline{1,q_{i}}\) ~ \(\forall i\))
 (instead of \(c_{ij}\));
 (2) an aggregated multiset estimate as the ``generalized median''
 (or ``set median'')
 is used for the objective function (``maximization'').
 The  item set is:
  ~ \(A= \bigcup_{i=1}^{m} A_{i}\),
 ~\( A_{i} = \{ (i,1),(i,2),...,(i,q_{i})\} \).
 The solution  is a subset of the initial item set:
 \( S = \{ (i,j) | x_{i,j}=1 \} \).
 Formally,
%
 \[ \max~ e(S) =   \max~ M = ~
  \arg \min_{M \in \overline{D} } ~~
  \sum_{(i,j) \in S=\{(i,j)| x_{i,j}=1\}} ~ | \delta (M, e_{i,j}) |\]
 \[ s.t. ~ \sum_{i=1}^{m} \sum_{j=1}^{q_{i}}  a_{ij} x_{i,j} \leq b,
 ~ \sum_{j=1}^{q_{i}} x_{ij} =  1,
 ~~~~ x_{ij} \in \{0,1\}.\]
 Evidently, this problem is similar to the above-mentioned combinatorial synthesis
 problem without compatibility of the selected items (objects, alternatives)
 \cite{lev12a,lev15}.
%

\subsubsection{Multiple knapsack problem with multiset estimates}

 The basic multiple knapsack problem is the following
 (e.g., \cite{chek05,gar79,hung78,keller04,mar90,pisinger99}):
%
 (i) item set \( A= \{ 1,...,i,...,m \} \);
 (ii) knapsack set
   \(B =  \{ B_{1},...,B_{j},...,B_{k} \}\) (\( k \leq m\));
 (iii) parameters   \( \forall i \in A \):
  profit \(c_{i}\),
 resource requirement (e.g., weight) \(a_{i}\); and
 (iv) resource (capacity) of knapsack \( B_{j} \in B \):
  \( b_{j}\).
 This problem is a special case of generalized assignment problem
 (multiple knapsack problem contains bin packing problem as special case).
 The model (i.e., ``\(0-1\) multiple knapsack problem'') is:
%
%
 \[\max  \sum_{j=1}^{k}  \sum_{i=1}^{m}  \gamma_{i} x_{ij}
%
 ~~~ s.t. ~  \sum_{i=1}^{m}   a_{i} x_{ij} \leq b_{j}, ~ \forall  j=\overline{1,k},
%
 ~  \sum_{j=1}^{k}   x_{ij} \leq 1 ,  \forall  i=\overline{1,m},
%
 ~ x_{ij} \in \{0,1\},  i=\overline{1,m},   j=\overline{1,k},\]
%
%
%
%
 where \(x_{ij}=1\) if item \(i\) is selected for knapsack \(B_{j}\),
  and \(x_{ij}=0\) otherwise.
%

%
 In the case of multiset estimates,
 item ``utility'' \(e_{i}, i=\overline{1,m}\)
 (instead of \(c_{i}\)) is considered.
%
%
 Multiple knapsack problem with multiset estimates
 and the integrated estimate for the solution is
 (solution \(S = \{ (i,j) | x_{ij} = 1\}\)):
%
 \[ \max~ e(S) = \biguplus_{ (i,j) \in S=\{ (i,j) | x_{i,j}=1\}} ~  e_{i},\]
 \[s.t. ~~~  \sum_{i=1}^{m}   a_{i} x_{ij} \leq b_{j}, ~ \forall  j=\overline{1,k},
%
 ~~~  \sum_{j=1}^{k}   x_{ij} \leq 1 , ~ \forall  i=\overline{1,m},
%
 ~~~~~ x_{ij} \in\{0,1\}, ~ i=\overline{1,m},  ~ j=\overline{1,k}.\]
%
%
%
%
%
%
 In the case of objective function based on median estimate
 for solution, the problem is:
%
%
 \[ \max~ e(S) =  \max~ M = arg \min_{M\in D} ~~
 | \biguplus_{ (i,j) \in S=\{ (i,j) | x_{ij}=1\}} ~   \delta (M,e_{i}) |\]
 \[s.t. ~~~  \sum_{i=1}^{m}   a_{i} x_{ij} \leq b_{j}, ~ \forall  j=\overline{1,k},
%
 ~~~  \sum_{j=1}^{k}   x_{ij} \leq 1 , ~ \forall  i=\overline{1,m},
%
 ~~~~~ x_{ij} \in\{0,1\}, ~ i=\overline{1,m}, ~  j=\overline{1,k}.\]
%
%

 In addition, it is reasonable to consider a new problem
 formulation while taking into account the number of the selected
 items (i.e., a special two-objective knapsack problem with multiset estimates)
 (solution \(S = \{ (i,j) | x_{ij} = 1\}\)):
%
%
 \[ \max~ e(S) =  \max~ M = arg \min_{M\in D} ~~
 | \biguplus_{ (i,j) \in S=\{ (i,j) | x_{i,j}=1\}} ~   \delta (M,e_{i}) |\]
 \[\\max \sum_{i=1}^{n} x_{i,j}   \]
 \[s.t. ~~~  \sum_{i=1}^{m}   a_{i} x_{ij} \leq b_{j}, ~ \forall  j=\overline{1,k}.
%
 ~~~  \sum_{j=1}^{k}   x_{ij} \leq 1 , ~ \forall  i=\overline{1,m},
 ~~~~~ x_{ij} \in\{0,1\}, ~ i=\overline{1,m},  ~ j=\overline{1,k}.\]
 Here, ``two''-dimensional
 space of solution quality (Fig. 9)
 can be considered as well.

\subsubsection{Assignment and generalized assignment problems with multiset estimates}

 The basic assignment problem is the following
 (e.g., \cite{gar79,shm93}).
 Simple {\it assignment problem} involves nonnegative correspondence matrix
 ~\( \Upsilon = || \gamma_{ij} ||\) ~(\(i=\overline{1,m}\),
 \(j=\overline{1,m}\))~
 where \(c_{ij}\) is a profit ('utility') to assign
 element \(i\) to position \(j\).
 The problem is (e.g., \cite{gar79}):

~~

     {\it Find the assignment} ~\(\pi =(\pi(1),...,\pi(m))\)~
     {\it of elements}
    \(i\) (\(i=\overline{1,m}\))
      {\it to positions}
      ~\(\pi(i)\)
      {\it which}
     {\it corresponds to a total effectiveness:}~
     \(\sum_{i=1}^{m} \gamma_{i\pi(i)} \rightarrow \max\).

~~

 The simplest algebraic problem formulation is:
 \[ \max \sum_{i=1}^{m} \sum_{j=1}^{m} \gamma_{i,j} x_{i,j}
%
 ~~~~ s.t. ~ \sum_{i=1}^{m} x_{i,j} \leq 1, ~ j=\overline{1,m};
 ~ \sum_{j=1}^{m} x_{i,j} = 1, ~ i=\overline{1,m};
 ~ x_{i,j} \in \{0,1\},  i=\overline{1,m},  j=\overline{1,m}. \]
 Here \(x_{i,j}=1\) if element \(i\) is assigned into
 position \(j\),
  \(c_{ij}\) is a profit (``utility'') of this assignment.
 The problem can be solved efficiently,
 for example, on the basis of Hungarian method (e.g., \cite{kuh57}).
 Note this problem is the matching problem for a bipartite graph
 (e.g., \cite{gar79}).

 In the generalized assignment problem,
 each item \(i\)
 (\(i=\overline{1,m}\))
  can be assigned to \(k\) (\(k \leq m\)) positions
 (knapsacks, bins) and
 a capacity is considered for each position \(j\) (\(j=\overline{1,k}\))
 (with corresponding capacity constraint
 \( \leq b_{j}\)) (Fig. 10).

\begin{center}
\begin{picture}(65,44.5)
\put(00,00){\makebox(0,0)[bl]{Fig. 10.
  Generalized assignment problem}}

\put(01.5,40){\makebox(0,8)[bl]{Items}}

\put(06,22.5){\oval(12,32)}

\put(06,35){\circle*{1.7}} \put(06,30){\circle*{1.7}}
\put(06,25){\circle*{1.7}} \put(06,20){\circle*{1.7}}
\put(06,15){\circle*{1.7}} \put(06,10){\circle*{1.7}}

\put(14,35){\vector(4,-1){18}}

\put(14,30){\vector(1,0){18}}

\put(14,25){\vector(1,0){18}}


\put(14,20){\vector(1,0){18}}

\put(14,15){\vector(1,0){18}}

\put(14,10){\vector(4,1){18}}

\put(33,39.5){\makebox(0,8)[bl]{Positions (e.g., bins,}}
\put(39,36){\makebox(0,8)[bl]{knapsacks)}}

\put(48,22.5){\oval(30,24)}

\put(35,30){\circle{2}} \put(35,30){\circle*{1}}
\put(35,30){\line(1,0){5}}

\put(40,28){\line(1,0){20}} \put(40,32){\line(1,0){20}}
\put(40,28){\line(0,1){4}} \put(60,28){\line(0,1){4}}

\put(35,25){\circle{2}} \put(35,25){\circle*{1}}
\put(35,25){\line(1,0){5}}

\put(40,23){\line(1,0){10}} \put(40,27){\line(1,0){10}}
\put(40,23){\line(0,1){4}} \put(50,23){\line(0,1){4}}

\put(35,20){\circle{2}} \put(35,20){\circle*{1}}
\put(35,20){\line(1,0){5}}

\put(40,18){\line(1,0){14}} \put(40,22){\line(1,0){14}}
\put(40,18){\line(0,1){4}} \put(54,18){\line(0,1){4}}

\put(35,15){\circle{2}} \put(35,15){\circle*{1}}
\put(35,15){\line(1,0){5}}

\put(40,13){\line(1,0){18}} \put(40,17){\line(1,0){18}}
\put(40,13){\line(0,1){4}} \put(58,13){\line(0,1){4}}

\end{picture}
\end{center}

 Formally,
 \[ \max \sum_{i=1}^{m} \sum_{j=1}^{k} \gamma_{i,j} x_{i,j}
%
 ~~~~ s.t. ~~ \sum_{i=1}^{m} x_{i,j} \leq 1, ~ j=\overline{1,k};
 ~ \sum_{j=1}^{k} x_{i,j} \geq 1, ~ i=\overline{1,m};
 ~ x_{i,j} \in \{0,1\},  i=\overline{1,m},  j=\overline{1,k}. \]
 In the case of multiset estimates,
 item ``utility''
 ~ \(e_{ij}\)
  ~ (\(i=\overline{1,m}\) ~ \(j=\overline{1,k}\))
 instead of \(c_{ij}\) is considered.
%

%
 The generalized assignment problem with multiset estimates
 and the integrated estimate for the solution is
 (solution \(S = \{ (i,j) | x_{ij} = 1\}\)):
%
 \[ \max~ e(S) = \biguplus_{ (i,j) \in S=\{ (i,j) | x_{i,j}=1\}} ~  e_{i},\]
 \[s.t. ~~~  \sum_{i=1}^{m}   a_{i} x_{ij} \leq b_{j}, ~ \forall  j=\overline{1,k},
%
 ~~~  \sum_{j=1}^{k}   x_{ij} = 1 , ~ \forall  i=\overline{1,m},
%
 ~~~~~ x_{ij} \in\{0,1\}, ~ i=\overline{1,m},  j=\overline{1,k}.\]
%
%
%
%
%
%

%
 In the case of objective function based on median estimate
 for solution, the problem is:
%
%
 \[ \max~ e(S) =  \max~ M = arg \min_{M\in D} ~~
 | \biguplus_{ (i,j) \in S=\{ (i,j) | x_{ij}=1\}} ~   \delta (M,e_{i}) |\]
 \[s.t. ~~~  \sum_{i=1}^{m}   a_{i} x_{ij} \leq b_{j}, ~ \forall  j=\overline{1,k},
%
 ~~~  \sum_{j=1}^{k}   x_{ij} = 1, ~ \forall  i=\overline{1,m},
%
 ~~~~~ x_{ij} \in\{0,1\}, ~ i=\overline{1,m},   j=\overline{1,k}.\]
%
%

 In addition, it is reasonable to consider a new problem
 formulation while taking into account the number of the selected
 items (i.e. a special two-objective generalized assignment problem with multiset estimates)
 (solution \(S = \{ (i,j) | x_{ij} = 1\}\)):
%
%
%
 \[ \max~ e(S) =  \max~ M = arg \min_{M\in D} ~~
 | \biguplus_{ (i,j) \in S=\{ (i,j) | x_{i,j}=1\}} ~   \delta (M,e_{i}) |
%
%
 ~~~~~~~~~ \max~ \sum_{i=1}^{n} x_{i,j} \]
 \[s.t. ~~~  \sum_{i=1}^{m}   a_{i} x_{ij} \leq b_{j}, ~ \forall  j=\overline{1,k}.
%
 ~~~  \sum_{j=1}^{k}   x_{ij} = 1, ~ \forall  i=\overline{1,m},
 ~~~~~ x_{ij} \in\{0,1\}, ~ i=\overline{1,m},   j=\overline{1,k}.\]
 Here, ``two''-dimensional space of solution quality (Fig. 9)
 can be considered as well.

\subsection{Inverse bin packing problem with multiset estimates}

 Generally, the inverse bin packing problem can be formulated
 as multiple knapsack problem with equal knapsack (i.e., bins).

 First, the basic inverse bin packing problem
 (with maximization of packed items),
 i.e., maximum cardinality bin packing problem,
 is considered as follows
 (e.g., \cite{ass84,bruno85,cof78,chung12,eps03,labbe95,labbe03,peet06}).
 Problem components are:
%
%
 (i) item set \( A= \{ 1,...,i,...,m \} \);
 (ii)  set of equal bins
   \(B =  \{ B_{1},...,B_{j},...,B_{k} \}\)
   (usually, \( k \leq m\));
 (iii) parameters   \( \forall i \in A \):
  profit \(\gamma_{i}\),
 resource requirement (e.g., weight) \(a_{i}\); and
 (iv) equal resource (capacity) of each bin \( B_{j} \in B \):
  \( b\).
 The model is:
 \[\max  \sum_{j=1}^{k}  \sum_{i=1}^{m}  \gamma_{i} x_{ij}
%
 ~~~~ s.t. ~  \sum_{i=1}^{m}   a_{i} x_{ij} \leq b,  \forall  j=\overline{1,k},
%
 ~  \sum_{j=1}^{k}   x_{ij} \leq 1, ~ \forall  i=\overline{1,m},
%
 ~ x_{ij} \in \{0,1\}, ~i=\overline{1,m},   ~j=\overline{1,k},\]
%
%
%
 where \(x_{ij}=1\) if item \(i\) is selected for knapsack \(B_{j}\),
  and \(x_{ij}=0\) otherwise.

 In the case of multiset estimates,
 item ``utility'' \(e_{i}, i=\overline{1,m}\)
 (instead of \(c_{i}\)) is considered.
 The inverse bin packing problem with multiset estimates
 and the integrated estimate for the solution is
 (solution \(S = \{ (i,j) | x_{ij} = 1\}\)):
%
 \[ \max~ e(S) = \biguplus_{ (i,j) \in S=\{ (i,j) | x_{i,j}=1\}} ~  e_{i},\]
 \[s.t. ~~~  \sum_{i=1}^{m}   a_{i} x_{ij} \leq b, ~ \forall  j=\overline{1,k},
%
 ~~~  \sum_{j=1}^{k}   x_{ij} \leq 1 , ~ \forall  i=\overline{1,m},
%
 ~~~~~ x_{ij} \in\{0,1\}, ~ i=\overline{1,m},  ~ j=\overline{1,k}.\]
%
%
%
%
%
%
 In the case of objective function based on median estimate
 for solution, the problem is:
%
%
 \[ \max~ e(S) =  \max~ M = arg \min_{M\in D} ~~
 | \biguplus_{ (i,j) \in S=\{ (i,j) | x_{ij}=1\}} ~   \delta (M,e_{i}) |\]
 \[s.t. ~~~  \sum_{i=1}^{m}   a_{i} x_{ij} \leq b, ~ \forall  j=\overline{1,k},
%
 ~~~  \sum_{j=1}^{k}   x_{ij} \leq 1 , ~ \forall  i=\overline{1,m},
%
 ~~~~~ x_{ij} \in\{0,1\}, ~ i=\overline{1,m}, ~  j=\overline{1,k}.\]
%
%
%
 The problem formulation while taking into account the number of the selected
 items (i.e., a special two-objective inverse bin packing problem with multiset estimates)
 (solution \(S = \{ (i,j) | x_{ij} = 1\}\)) is:
%
%
%
 \[ \max~ e(S) = \max~ M = arg \min_{M\in D} ~~
 | \biguplus_{ (i,j) \in S=\{ (i,j) | x_{i,j}=1\}} ~ \delta (M,e_{i}) |
%
%
 ~~~~~~~~ \max~ \sum_{i=1}^{n}~ x_{i,j}\]
 \[s.t. ~~~  \sum_{i=1}^{m}~  a_{i} x_{ij} \leq b, ~ \forall  j=\overline{1,k}.
%
 ~~~  \sum_{j=1}^{k}~   x_{ij} \leq 1 , ~ \forall  i=\overline{1,m},
 ~~~~~ x_{ij} \in\{0,1\}, ~ i=\overline{1,m},  ~ j=\overline{1,k}.\]
 Here,  ``two''-dimensional space of solution quality (Fig. 9)
 can be considered as well.

\subsection{Bin packing with conflicts}

 The bin packing problem with conflict
 consists in packing items into the minimum number of bins
 subject to incompatibility constraints.
 (e.g., \cite{eps08,fern10,gend04,jans99,sad13}).
 The description of the problem is the following.
 Given a set of \(n\) items \(A\),
  corresponding their weights
 \(w_{1}\), \(w_{2}\),  ... \(w_{n}\),
  and a set of identical bins
  (\(k = 1,2,...\)) with capacity \(b\).
 It can be assumed: \(w_{1} \geq w_{2} \geq ... \geq w_{n}\).
%
 Given
 conflict relation over items as conflict graph
 \(G = (A,E)\), where an edge \( (\iota_{1},\iota_{2}) \in  E\) exists
 if and only if items \(\iota_{1}, \iota_{2} \in A\) conflict
 or \( w_{\iota_{1}} + w_{\iota_{2}} > b \).
 Let \(y_{k}\) be a binary variable:
 \(y_{k} = 1\)  if bin \(k\) is used,
 and \(x_{\iota k}\) be a binary variable:
 \(x_{\iota k} = 1\)  if item \(i\) is assigned to bin \(k\).
 Formally,
 \[min~ z = \sum_{\iota =1}^{n} ~y_{k} \]
 \[s.t. ~~~~ \sum_{ \iota =1}^{n} ~w_{\iota} x_{ \iota k} \leq b y_{k} ~\forall k=\overline{1,n};
 ~~ \sum_{ \iota =1}^{n} ~x_{\iota k} = 1 ~\forall \iota=\overline{1,n};
 ~~ x_{\iota_{2} k} +  x_{\iota_{2} k} \leq 1 ~\forall (\iota_{1} ,\iota_{2}) \in E, ~\forall k=\overline{1,n};   \]
 \[  y_{k} \in \{0,1\} ~\forall k=\overline{1,n}; ~~  x_{\iota k} \in \{0,1\} ~
 \forall \iota =\overline{1,n}, ~\forall k=\overline{1,n}. \]
 Evidently, the problem generalizes the classic bin packing problem and is HP-hard
 (e.g., \cite{mar90}).

 In inverse bin packing problem
 (maximization of the number of packed items subject to fixed set of bins),
 The problem is as follows.
 Let \(\gamma_{\iota}\) be an importance (utility, profit)
 of packing item \(\iota \in A\).
 Formally,
 \[max~ \sum_{\iota =1}^{n}  \sum_{k=1}^{q}  \gamma_{\iota}    ~x_{\iota k} \]
 \[s.t. ~~~~ \sum_{ \iota =1}^{n} ~w_{\iota} x_{ \iota k} \leq b, ~\forall k=\overline{1,n};
 ~~ \sum_{ \iota =1}^{n} ~x_{\iota k} \leq 1 ~\forall \iota=\overline{1,n};
 ~~ x_{\iota_{2} k} +  x_{\iota_{2} k} \leq 1 ~~ (\iota_{1} ,\iota_{2}) \in E, ~\forall k=\overline{1,n};\]
 \[  x_{\iota k} \in \{0,1\} ~
 \forall \iota =\overline{1,n}, ~\forall k=\overline{1,n}. \]
%
%
%
 Let \(e_{\iota}\) be  an importance  multiset estimate (utility, profit)
 of packing item \(\iota \in A\).
 The inverse bin packing problem with multiset estimates
 and the integrated estimate for the solution is
 (solution \(S = \{ (\iota ,k) | x_{\iota k} = 1\}\)):
%
%
 \[ \max~ e(S) = \biguplus_{ (\iota ,k) \in S=\{ (\iota ,k) | x_{\iota ,k}=1\}} ~  e_{\iota },\]
%
%
%
 \[s.t. ~~~~ \sum_{ \iota =1}^{n} \sum_{k=1}^{q}   ~w_{\iota,k} x_{ \iota k} \leq b, ~\forall k=\overline{1,n};
 ~~ \sum_{ \iota =1}^{n} ~x_{\iota, k} \leq 1 ~\forall \iota=\overline{1,n};
 ~~ x_{\iota_{2} k} +  x_{\iota_{2} k} \leq 1 ~~ (\iota_{1} ,\iota_{2}) \in E, ~\forall k=\overline{1,n};\]
 \[  x_{\iota k} \in \{0,1\} ~\forall \iota =\overline{1,n}, ~\forall k=\overline{1,n}. \]

 In addition,
 objective function can be examine:
\[\max~ \sum_{\iota=1}^{n} \sum_{k=1}^{q}  x_{\iota ,k} ~
  ~\forall \iota =\overline{1,n}, ~\forall k=\overline{1,n}.\]

\newpage
\section{Colored bin packing}

\subsection{Basic colored bin packing}

 Now consider the basic colored bin packing problem
 (e.g., \cite{chungf06,twigl15}).
 A set of items
 \(A = \{a_{1},...,a_{i},...,a_{n}\}\)
  of different sizes
  (e.g., \(w_{i} \in (0,1]\) ~\(\forall i=\overline{1,n}\))
  is given.
  It is necessary to pack the items above into bins of
  equal size
   so that
 a few bins is used in total (at most \(\alpha\) times optimal),
 and that the items of each color span few bins
  (at most \(\beta\) times optimal).
 The obtained allocations are called \(\alpha,\beta\)-approximate.
 The colored bin packing problem corresponds to many significant applications,
 for example (e.g., \cite{twigl15}):
 (1) allocating files in P2P networks,
 (2) allocating related jobs (i.e., related jobs are of the same color)
 to processors,
 (3) allocating related items in a distributed cache,
 and
  (4) allocating jobs in a grid computing system.
 Fig. 11 illustrates the colored bin packing problem:
 eleven items, three  colors (\(\lambda\), \(\mu\), \(\theta\)).
 The illustrative solution is:
 (i)  color \(\lambda\) for bin \(1\), bin \(2\);
 (ii)  color \(\mu\) for bin \(3\); and
 (iii)  color \(\theta\) for bin \(4\), bin \(5\).

\begin{center}
\begin{picture}(125,78)
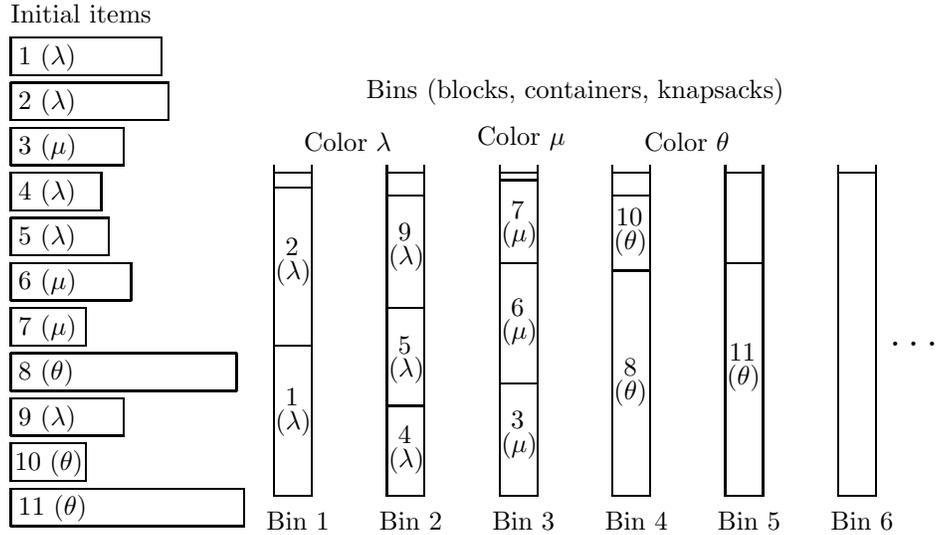
\put(25,00){\makebox(0,0)[bl]{Fig. 11. Illustration for colored
 bin-packing}}

\put(00,73){\makebox(0,0)[bl]{Initial items}}

\put(00,66){\line(1,0){20}} \put(00,71){\line(1,0){20}}
\put(00,66){\line(0,1){05}} \put(20,66){\line(0,1){05}}

\put(01,66.6){\makebox(0,0)[bl]{1 (\(\lambda\))}}

\put(00,60){\line(1,0){21}} \put(00,65){\line(1,0){21}}
\put(00,60){\line(0,1){05}} \put(21,60){\line(0,1){05}}

\put(01,60.6){\makebox(0,0)[bl]{2 (\(\lambda\))}}

\put(00,54){\line(1,0){15}} \put(00,59){\line(1,0){15}}
\put(00,54){\line(0,1){05}} \put(15,54){\line(0,1){05}}

\put(01,54.6){\makebox(0,0)[bl]{3 (\(\mu\))}}

\put(00,48){\line(1,0){12}} \put(00,53){\line(1,0){12}}
\put(00,48){\line(0,1){05}} \put(12,48){\line(0,1){05}}

\put(01,48.6){\makebox(0,0)[bl]{4 (\(\lambda\))}}

\put(00,42){\line(1,0){13}} \put(00,47){\line(1,0){13}}
\put(00,42){\line(0,1){05}} \put(13,42){\line(0,1){05}}

\put(01,42.6){\makebox(0,0)[bl]{5 (\(\lambda\))}}

\put(00,36){\line(1,0){16}} \put(00,41){\line(1,0){16}}
\put(00,36){\line(0,1){05}} \put(16,36){\line(0,1){05}}

\put(01,36.6){\makebox(0,0)[bl]{6 (\(\mu\))}}

\put(00,30){\line(1,0){10}} \put(00,35){\line(1,0){10}}
\put(00,30){\line(0,1){05}} \put(10,30){\line(0,1){05}}

\put(01,30.6){\makebox(0,0)[bl]{7 (\(\mu\))}}

\put(00,24){\line(1,0){30}} \put(00,29){\line(1,0){30}}
\put(00,24){\line(0,1){05}} \put(30,24){\line(0,1){05}}

\put(01,24.6){\makebox(0,0)[bl]{8 (\(\theta\))}}

\put(00,18){\line(1,0){15}} \put(00,23){\line(1,0){15}}
\put(00,18){\line(0,1){05}} \put(15,18){\line(0,1){05}}

\put(01,18.6){\makebox(0,0)[bl]{9 (\(\lambda\))}}

\put(00,12){\line(1,0){10}} \put(00,17){\line(1,0){10}}
\put(00,12){\line(0,1){05}} \put(10,12){\line(0,1){05}}

\put(0.5,12.6){\makebox(0,0)[bl]{10 (\(\theta\))}}

\put(00,06){\line(1,0){31}} \put(00,11){\line(1,0){31}}
\put(00,06){\line(0,1){05}} \put(31,06){\line(0,1){05}}

\put(01,06.6){\makebox(0,0)[bl]{11 (\(\theta\))}}

\put(47.3,62){\makebox(0,0)[bl]{Bins (blocks, containers,
 knapsacks)}}

\put(39,56){\makebox(0,0)[bl]{Color  \(\lambda\)}}

\put(62,56){\makebox(0,0)[bl]{Color  \(\mu\)}}

\put(84.2,56){\makebox(0,0)[bl]{Color  \(\theta\)}}


\put(35,10){\line(1,0){05}} \put(35,53){\line(1,0){05}}
\put(35,10){\line(0,1){44}} \put(40,10){\line(0,1){44}}

\put(34,05.5){\makebox(0,0)[bl]{Bin 1}}

\put(35,30){\line(1,0){05}} \put(36.5,22){\makebox(0,0)[bl]{1}}
\put(35.2,18){\makebox(0,0)[bl]{(\(\lambda\))}}

\put(35,51){\line(1,0){05}} \put(36.5,42){\makebox(0,0)[bl]{2}}
\put(35.2,38){\makebox(0,0)[bl]{(\(\lambda\))}}

\put(50,10){\line(1,0){05}} \put(50,53){\line(1,0){05}}
\put(50,10){\line(0,1){44}} \put(55,10){\line(0,1){44}}

\put(49,05.5){\makebox(0,0)[bl]{Bin 2}}

\put(50,22){\line(1,0){05}} \put(51.5,17){\makebox(0,0)[bl]{4}}
\put(50.2,13){\makebox(0,0)[bl]{(\(\lambda\))}}

\put(50,35){\line(1,0){05}} \put(51.5,29){\makebox(0,0)[bl]{5}}
\put(50.2,25){\makebox(0,0)[bl]{(\(\lambda\))}}

\put(50,50){\line(1,0){05}} \put(51.5,44){\makebox(0,0)[bl]{9}}
\put(50.2,40){\makebox(0,0)[bl]{(\(\lambda\))}}

\put(65,10){\line(1,0){05}} \put(65,53){\line(1,0){05}}
\put(65,10){\line(0,1){44}} \put(70,10){\line(0,1){44}}

\put(64,05.5){\makebox(0,0)[bl]{Bin 3}}

\put(65,25){\line(1,0){05}} \put(66.5,19){\makebox(0,0)[bl]{3}}
\put(65.2,15){\makebox(0,0)[bl]{(\(\mu\))}}

\put(65,41){\line(1,0){05}} \put(66.5,34){\makebox(0,0)[bl]{6}}
\put(65.2,30){\makebox(0,0)[bl]{(\(\mu\))}}

\put(65,52){\line(1,0){05}} \put(66.5,47){\makebox(0,0)[bl]{7}}
\put(65.2,43){\makebox(0,0)[bl]{(\(\mu\))}}

\put(80,10){\line(1,0){05}} \put(80,53){\line(1,0){05}}
\put(80,10){\line(0,1){44}} \put(85,10){\line(0,1){44}}

\put(79,05.5){\makebox(0,0)[bl]{Bin 4}}

\put(80,40){\line(1,0){05}}
\put(81.5,26){\makebox(0,0)[bl]{8}}
\put(80.2,22){\makebox(0,0)[bl]{(\(\theta\))}}

\put(80,50){\line(1,0){05}}
\put(80.5,46){\makebox(0,0)[bl]{10}}
\put(80.2,42){\makebox(0,0)[bl]{(\(\theta\))}}

\put(95,10){\line(1,0){05}} \put(95,53){\line(1,0){05}}
\put(95,10){\line(0,1){44}} \put(100,10){\line(0,1){44}}

\put(94,05.5){\makebox(0,0)[bl]{Bin 5}}

\put(95,41){\line(1,0){05}}
\put(95.5,28){\makebox(0,0)[bl]{11}}
\put(95.2,24){\makebox(0,0)[bl]{(\(\theta\))}}

\put(110,10){\line(1,0){05}} \put(110,53){\line(1,0){05}}
\put(110,10){\line(0,1){44}} \put(115,10){\line(0,1){44}}

\put(109,05.5){\makebox(0,0)[bl]{Bin 6}}

\put(117,30){\makebox(0,0)[bl]{{\bf .~.~.}}}


\end{picture}
\end{center}
%

 Recently, some  versions of colored bin packing problem
 have been examined:
 (1) basic colored bin packing \cite{chungf06,twigl15},
 (2) offline colored bin packing \cite{twigl15},
 (3) online colored bin packing \cite{twigl15},
 and
 (4) online bin coloring (packing with minimum colors)
 \cite{krumke01}.

\subsection{Two auxiliary graph coloring problems}

\subsubsection{Auxiliary vertex graph coloring problem with ordinal color proximity}

 First,  the vertex coloring problem is considered as a basic one.
  The problem can be described as the following
  (e.g.,
  \cite{camp04,ensen95,gar79,harary69,harary85,kubal04,nemh88,west01}).
 Given undirected graph
  \( G=(A,E) \)
  (a node/vertices set \(A\) and
  an edge set \(E\), \(|A| = n\)).
  There is a set of colors (labels, numbers)
  \(X = \{ x_{1},...,x_{l},...,x_{k} \} \).
 Let ~\(C(G) = \{ C(a_{1}),...,C(a_{i}),...,C(a_{i}) \} \)
 (\(C_{a_{i}} \in X\))
 (or  \( < C(a_{1}) \star ... \star C(a_{i})\star ... \star C(a_{i}) > \) )
 be a {\it color configuration}
 (i.e., assignment of a color for each vertex).
 The problem is:

~~

 {\it Assign for each vertex} ~\( \forall a_{i} \in A \)
 {\it  label or color}   ~({\it i.e.,} \( C(a_{i}) \))
  {\it such that no edge connects two identical colored vertices,
 i.e.,}
  ~\(\forall a_{i},a_{j} \in A\)
 ~~{\it if} ~\( (a_{i},a_{j}) \in E \) ({\it i.e., adjacent vertices})
 ~~{\it then} \( C(a_{i}) \neq C(a_{j}) \).

~~

 Thus,  {\it color configuration}
 (e.g., ~\( C(G) = \{ C(a_{1}),...,C(a_{i}),...,C(a_{n}) \} \))
 for a given graph \(G=(A,E)\)
 is searched for.
 Clearly, \(|C(G)|\) equals the number of used colors (labels).
 (The minimal number of required colors for a graph \(G\)
 is called {\it chromatic number} of the graph \( \chi(G)\)).
 Note, other coloring problems
 can be transformed into the vertex
 version.
 Fig. 12 illustrates the vertex coloring problem:
 ~\(G = (A,E), ~A = \{p,q,u,v,w\}, ~E=\{(p,q),(p,u),(q,v),(u.v),(w,p)(w,q)(w,u)(w,v)\}\)
 and three colors \(\{ x_{1},x_{2},x_{3}\}\)
 (i.e., corresponding indices for
 colors of vertices).

\begin{center}
\begin{picture}(60,42)
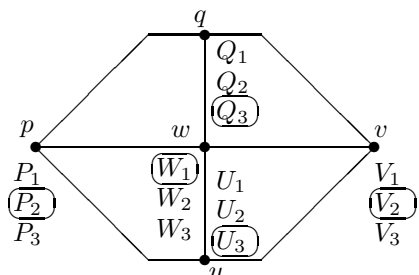
\put(00,00){\makebox(0,0)[bl]{Fig. 12. Example of
 vertex coloring }}
%
\put(05,22){\circle*{1.5}}
\put(27.5,37){\circle*{1.5}}
\put(27.5,07){\circle*{1.5}}
\put(50,22){\circle*{1.5}}
\put(27.5,22){\circle*{1.5}}

\put(05,22){\line(1,-1){15}} \put(20,07){\line(1,0){15}}

\put(05,22){\line(1,0){45}} \put(05,22){\line(1,1){15}}

\put(20,37){\line(1,0){15}}

\put(27.5,37){\line(0,-1){15}} \put(27.5,22){\line(0,-1){15}}

\put(35,07){\line(1,1){15}}

\put(35,37){\line(1,-1){15}}
\put(03,23.4){\makebox(0,0)[bl]{\(p\)}}

\put(02,17){\makebox(0,0)[bl]{\(P_{1}\)}}
\put(02,13){\makebox(0,0)[bl]{\(P_{2}\)}}
\put(02,09){\makebox(0,0)[bl]{\(P_{3}\)}}
\put(04.5,14.5){\oval(06,4)}
%
\put(50,17){\makebox(0,0)[bl]{\(V_{1}\)}}
\put(50,13){\makebox(0,0)[bl]{\(V_{2}\)}}
\put(50,09){\makebox(0,0)[bl]{\(V_{3}\)}}

\put(50,23.4){\makebox(0,0)[bl]{\(v\)}}
\put(52.5,14.5){\oval(06,4)}
%
\put(29,16){\makebox(0,0)[bl]{\(U_{1}\)}}
\put(29,12){\makebox(0,0)[bl]{\(U_{2}\)}}
\put(29,08){\makebox(0,0)[bl]{\(U_{3}\)}}

\put(28,04.5){\makebox(0,0)[bl]{\(u\)}}
\put(31.5,09.5){\oval(06,4)}
%

\put(29,33){\makebox(0,0)[bl]{\(Q_{1}\)}}
\put(29,29){\makebox(0,0)[bl]{\(Q_{2}\)}}
\put(29,25){\makebox(0,0)[bl]{\(Q_{3}\)}}

\put(26,38){\makebox(0,0)[bl]{\(q\)}}

\put(31.5,26.8){\oval(06,4)}
%
\put(21,17.6){\makebox(0,0)[bl]{\(W_{1}\)}}
\put(21,13.6){\makebox(0,0)[bl]{\(W_{2}\)}}
\put(21,09.6){\makebox(0,0)[bl]{\(W_{3}\)}}

\put(23,23.4){\makebox(0,0)[bl]{\(w\)}}
\put(23.5,19.1){\oval(06,4)}
%
%
\end{picture}
\end{center}

 The resultant {\it color configuration} (solution) is:
  ~\(C(G) = \{P_{2},W_{1},V_{2},Q_{3},U_{3}\}\).
 The number of possible resultant {\it color configurations}
 (three colors) equals \(6\):

 (1) \(C^{1}(G) = \{P_{1},W_{2},V_{1},Q_{3},U_{3}\}\),
 (2) \(C^{2}(G) = \{P_{1},W_{3},V_{1},Q_{2},U_{2}\}\),

 (3) \(C^{3}(G) = \{P_{3},W_{2},V_{3},Q_{1},U_{1}\}\),
 (4) \(C^{4}(G) = \{P_{3},W_{1},V_{3},Q_{2},U_{2}\}\),

 (5) \(C^{5}(G) = \{P_{2},W_{1},V_{2},Q_{3},U_{3}\}\),
 (6) \(C^{6}(G) = \{P_{2},W_{3},V_{2},Q_{1},U_{1}\}\).

 In addition,
 an aggregated weight (e.g., additive function) of used colors
 (each color has its nonnegative weight
 ~\( w(x_{l}) ~\forall x_{l} \in X, ~l=\overline{1,k} \)) can be considered as well.
 As a result, the following minimization problem formulation can be examined:

~~~~

 {\it Find  coloring of vertices}
 ~\(C^{*}(G)\)~
 {\it for a given graph} ~\(G=(A,E)\)
 {\it with the minimum number of used colors (labels)}:
%
%
 \[ \min_{\{C(G)\}} ~~| C^{*}( G=(A,E)) |
 ~~~~~s.t. ~~ C^{*}(a_{i}) \neq  C^{*}(a_{j}) ~~ \forall (a_{i},a_{j}) \in E, ~i \neq j.\]

~~

 This problem is NP-hard
 (e.g.,
  \cite{ensen95,gar79,johnson96,west01}).
 Let ~\( \overline{C^{*}(G)} = \{ c^{*}_{\theta} \}\)~ be the set of used colors
 ~(i.e., \(  \overline{C^{*}(G)}  \subseteq  C^{*}(G) \)).
 In the case of weighted colors (and additive aggregation
 function),
 the following model can be considered:
 \[ \min    \sum_{\forall c^{*}_{\theta} \in  \overline{C^{*}(G)}} ~~w(c^{*}_{\theta})
 ~~~~~~s.t. ~~ C^{*}(a_{i}) \neq  C^{*}(a_{j}) ~~ \forall (a_{i},a_{j}) \in E, ~i \neq j.\]
 Clearly, if
 ~\(w(x_{l} ) = 1 ~\forall x_{l} \in X\)~
  this problem formulation
 is equivalent to the previous one.
 In the case of vector-like color weight
  \[ (~ w^{1}(c_{\theta}),...,w^{\mu }(c_{\theta}),...,w^{\lambda }(c_{\theta})
  ~)~
 ~~\forall c_{\theta} \in C \]~
 and additive aggregation functions,
 the objective vector function is:
\[ (~ \sum_{\forall c^{*}_{\theta} \in  \overline{ C^{*}(G)}} ~~w^{1}(c^{*}_{\theta}) ~,...,
   \sum_{\forall c^{*}_{\theta} \in  \overline{ C^{*}(G)} } ~~w^{\mu }(c^{*}_{\theta}) ~,...,
   \sum_{\forall c^{*}_{\theta} \in \overline{ C^{*}(G)}} ~~w^{\lambda }(c^{*}_{\theta}) ~~~) \]
 and Pareto-efficient solutions
 by the vector function are searched for.

 Generally,
 it may be prospective to consider a set of objective functions
 (criteria) as follows
 (e.g., \cite{lev98,lev15}):
%
 ~(i) number of used colors,
 ~(ii) an aggregated weight of used colors,
 ~(iii) correspondence of colors to vertices
 (e.g., the worst correspondence, average correspondence)
  (e.g., \cite{lev98});
 ~(iv) quality of compatibility of colors,
 which were assigned to the neighbor (i.e., adjacent) vertices
  (e.g., the worst case, average case)
   (e.g., \cite{lev98}); and
 ~(v) conditions at a distance that equals three, four, etc.


 The author's version of graph (vertex) coloring problem
 (while taking into account color compatibility
 and correspondence of colors to vertices)
 is described in \cite{lev98} (numerical example, Fig. 13).
 Here,
 the solving approach is based on morphological clique problem
 (i.e., HMMD).
 Six colors are used:
  \(x_{1}\),
  \(x_{2}\),
  \(x_{3}\),
  \(x_{4}\),
  \(x_{5}\), and
  \(x_{6}\).
%
%
 Estimates of correspondence of colors to vertices are shown in
 parentheses in Fig. 13
 (\(1\) corresponds to the best level).
  Table 10  contains compatibility estimates  for colors
  (\(4\) corresponds to the best level).

\begin{center}
\begin{picture}(62,50)
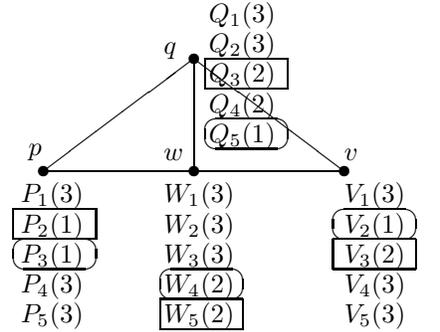
\put(00,00){\makebox(0,0)[bl]{{\bf Fig. 13.} Coloring with
 compatibility  }}

\put(05,26){\circle*{1.5}}

\put(25,41){\circle*{1.5}}


\put(45,26){\circle*{1.5}}

\put(25,26){\circle*{1.5}}
\put(05,26){\line(1,0){40}} \put(05,26){\line(4,3){20}}

\put(25,41){\line(0,-1){15}}


\put(45,26){\line(-4,3){20}}
\put(03,27.5){\makebox(0,0)[bl]{\(p\)}}

\put(02,21){\makebox(0,0)[bl]{\(P_{1}(3)\)}}
\put(02,17){\makebox(0,0)[bl]{\(P_{2}(1)\)}}
\put(02,13){\makebox(0,0)[bl]{\(P_{3}(1)\)}}
\put(02,09){\makebox(0,0)[bl]{\(P_{4}(3)\)}}
\put(02,05){\makebox(0,0)[bl]{\(P_{5}(3)\)}}
%
\put(06.5,15){\oval(11,4)}

\put(01,17){\line(1,0){11}} \put(01,21){\line(1,0){11}}
\put(01,17){\line(0,1){4}} \put(12,17){\line(0,1){4}}

%
\put(45,21){\makebox(0,0)[bl]{\(V_{1}(3)\)}}
\put(45,17){\makebox(0,0)[bl]{\(V_{2}(1)\)}}
\put(45,13){\makebox(0,0)[bl]{\(V_{3}(2)\)}}
\put(45,09){\makebox(0,0)[bl]{\(V_{4}(3)\)}}
\put(45,05){\makebox(0,0)[bl]{\(V_{5}(3)\)}}

\put(45,27.5){\makebox(0,0)[bl]{\(v\)}}
\put(49,19){\oval(11,4)}

\put(43.5,13){\line(1,0){11}} \put(43.5,17){\line(1,0){11}}
\put(43.5,13){\line(0,1){4}} \put(54.5,13){\line(0,1){4}}

%
%
\put(27,45){\makebox(0,0)[bl]{\(Q_{1}(3)\)}}
\put(27,41){\makebox(0,0)[bl]{\(Q_{2}(3)\)}}
\put(27,37){\makebox(0,0)[bl]{\(Q_{3}(2)\)}}
\put(27,33){\makebox(0,0)[bl]{\(Q_{4}(2)\)}}
\put(27,29){\makebox(0,0)[bl]{\(Q_{5}(1)\)}}

\put(21,41){\makebox(0,0)[bl]{\(q\)}}
\put(32,31){\oval(11,4)}

\put(26.5,37){\line(1,0){11}} \put(26.5,41){\line(1,0){11}}
\put(26.5,37){\line(0,1){4}} \put(37.5,37){\line(0,1){4}}

%
\put(21,21){\makebox(0,0)[bl]{\(W_{1}(3)\)}}
\put(21,17){\makebox(0,0)[bl]{\(W_{2}(3)\)}}
\put(21,13){\makebox(0,0)[bl]{\(W_{3}(3)\)}}
\put(21,09){\makebox(0,0)[bl]{\(W_{4}(2)\)}}
\put(21,05){\makebox(0,0)[bl]{\(W_{5}(2)\)}}

\put(21,27.5){\makebox(0,0)[bl]{\(w\)}}
\put(26,11){\oval(11,4)}

\put(20.5,05){\line(1,0){11}} \put(20.5,09){\line(1,0){11}}
\put(20.5,05){\line(0,1){4}} \put(31.5,05){\line(0,1){4}}

%
\end{picture}
\end{center}

  If the edge between vertices is absent
  the corresponding compatibility estimates of colors equal to the best
  level (i.e., \(4\) for vertex pair  ~\((p,v)\)).
 Two examples of color combinations (color compositions)
 and their quality vectors are the following (Fig. 14):

 (a) \(C^{*1}(G)=< P_{2} \star Q_{3} \star V_{3} \star  W_{5}  >\),
 ~\(N (C^{*1}(G)) = (4;1,3,0)\);

 (b)  \(C^{*2}(G)= <P_{3} \star Q_{5} \star V_{2} \star W_{4}>\),
 ~\(N (C^{*2}(G))=(2;3,1,0)\);

 (c)  \(C^{*3}(G)=<P_{2} \star Q_{5} \star V_{2} \star W_{5} >\),
 ~\(N (C^{*2}(G))=(2;3,1,0)\).

\begin{center}
{\bf Table 10.} Compatibility estimates of colors  \\
%
\begin{tabular}{|c|c c c c c  c c c c c  c c c c c |}
\hline

 &
 \(Q_{1}\)&\(Q_{2}\)&\(Q_{3}\)&\(Q_{4}\)&\(Q_{5}\)&

 \(V_{1}\)&\(V_{2}\)&\(V_{3}\)&\(V_{4}\)&\(V_{5}\)&

 \(W_{1}\)&\(W_{2}\)&\(W_{3}\)&\(W_{4}\)&\(W_{5}\)\\

\hline

 \(P_{1}\)&
 \(0\)&\(1\)&\(2\)&\(3\)&\(4\)&
 \(4\)&\(4\)&\(4\)&\(4\)&\(4\)&
 \(0\)&\(1\)&\(2\)&\(3\)&\(3\)\\

  \(P_{2}\)&
 \(1\)&\(0\)&\(4\)&\(2\)&\(3\)&
 \(4\)&\(4\)&\(4\)&\(4\)&\(4\)&
 \(1\)&\(0\)&\(1\)&\(2\)&\(4\)\\

  \(P_{3}\)&
 \(2\)&\(4\)&\(0\)&\(1\)&\(2\)&
 \(4\)&\(4\)&\(4\)&\(4\)&\(4\)&
 \(2\)&\(1\)&\(0\)&\(2\)&\(4\)\\

  \(P_{4}\)&
 \(3\)&\(2\)&\(1\)&\(0\)&\(3\)&
 \(4\)&\(4\)&\(4\)&\(4\)&\(4\)&
 \(3\)&\(2\)&\(1\)&\(0\)&\(2\)\\

  \(P_{5}\)&
 \(4\)&\(3\)&\(2\)&\(3\)&\(0\)&
 \(4\)&\(4\)&\(4\)&\(4\)&\(4\)&
 \(4\)&\(3\)&\(2\)&\(3\)&\(2\)\\

 \(Q_{1}\)&
   &&&&&
 \(0\)&\(1\)&\(2\)&\(3\)&\(4\)&
 \(0\)&\(1\)&\(2\)&\(3\)&\(3\)\\

 \(Q_{2}\)&  &&&&&
 \(1\)&\(0\)&\(4\)&\(2\)&\(3\)&
 \(1\)&\(0\)&\(1\)&\(2\)&\(4\)\\

 \(Q_{3}\)&  &&&&&
 \(4\)&\(1\)&\(4\)&\(1\)&\(2\)&
 \(2\)&\(1\)&\(0\)&\(1\)&\(4\)\\

 \(Q_{4}\)&  &&&&&
 \(3\)&\(2\)&\(1\)&\(0\)&\(3\)&
 \(3\)&\(2\)&\(1\)&\(0\)&\(2\)\\

 \(Q_{5}\)&  &&&&&
 \(4\)&\(3\)&\(2\)&\(3\)&\(0\)&
 \(4\)&\(3\)&\(2\)&\(3\)&\(3\)\\

 \(V_{1}\)&  &&&&& &&&&&
 \(0\)&\(1\)&\(2\)&\(3\)&\(3\)\\

 \(V_{2}\)&  &&&&& &&&&&
 \(1\)&\(0\)&\(1\)&\(2\)&\(4\)\\

 \(V_{3}\)&  &&&&& &&&&&
 \(2\)&\(1\)&\(0\)&\(1\)&\(4\)\\

 \(V_{4}\)&  &&&&& &&&&&
 \(3\)&\(2\)&\(1\)&\(0\)&\(2\)\\

 \(V_{5}\)&  &&&&& &&&&&
 \(4\)&\(3\)&\(2\)&\(3\)&\(2\)\\

\hline
\end{tabular}
\end{center}
%

\begin{center}
\begin{picture}(69,67)

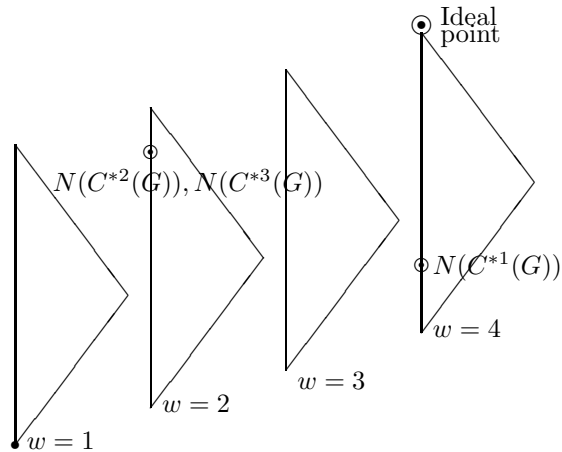
\put(00,00){\makebox(0,0)[bl]{Fig. 14.  Poset-like scale for color
 configuration}}

\put(00,06){\circle*{0.9}}

\put(00,06){\line(0,1){40}} \put(00,06){\line(3,4){15}}
\put(00,46){\line(3,-4){15}}

\put(18,11){\line(0,1){40}} \put(18,11){\line(3,4){15}}
\put(18,51){\line(3,-4){15}}

\put(36,16){\line(0,1){40}} \put(36,16){\line(3,4){15}}
\put(36,56){\line(3,-4){15}}

\put(54,21){\line(0,1){40}} \put(54,21){\line(3,4){15}}
\put(54,61){\line(3,-4){15}}


\put(18,45){\circle*{0.75}} \put(18,45){\circle{1.7}}
\put(05,39){\makebox(0,0)[bl]{\(N(C^{*2}(G)),N(C^{*3}(G))\) }}

\put(54,30){\circle*{0.75}} \put(54,30){\circle{1.7}}
\put(55.5,28){\makebox(0,0)[bl]{\(N(C^{*1}(G))\)}}


\put(54,62){\circle*{1}} \put(54,62){\circle{2.5}}

\put(56.5,62){\makebox(0,0)[bl]{Ideal}}
\put(56.5,59){\makebox(0,0)[bl]{point}}

\put(01.5,05.5){\makebox(0,0)[bl]{\(w=1\)}}
\put(19.5,10.5){\makebox(0,0)[bl]{\(w=2\)}}
\put(37.5,13.5){\makebox(0,0)[bl]{\(w=3\)}}
\put(55.5,20.5){\makebox(0,0)[bl]{\(w=4\)}}

\end{picture}
\end{center}

\subsubsection{Partition coloring problem}

 Here
 the partition coloring problem
 (i.e., selective graph clustering over clustered graph)
 is considered as a close auxiliary problem
 \cite{frota10,hos11,li00,noro06}.
 The problem formulation is as follows.
 Given a non-directed graph \(G = (V,E)\),
 where \(V\) is the set of vertices (nodes) and
 \(E\) is the set of edges.
 Let \(\{V_{1},V_{2},...,V_{q}\}\)
 be a partition of \(V\) into \(q\)
 subsets with \(V = \bigcup_{\iota =1}^{q} V_{\iota} \)
 and \(|V_{\iota_{1}} \bigcap V_{\iota_{2}} |= 0\)
  ~\(\forall \iota_{1},\iota_{2} = 1,2,...,q \) with
 \(\iota_{1} \neq \iota_{2}\).
 Clearly,
 \(V_{\iota}\) (\(\forall \iota = \overline{1,q}\)) is  a graph part or a graph component.
 The partition coloring problem is:

~~

 Find a subset \(V' \subseteq V\) such that
 \( | V' \bigcap V_{\iota} | = 1 \) \(\forall \iota = \overline{1,q}\)
 (i.e., \(V'\) contains one vertex from each component
 \(V_{\iota}\)),
 and the chromatic number of the graph induced in \(G\)
 by \(V'\) is minimum.

~~

 Evidently, the problem is a generalization of the graph coloring problem
  and belongs to class of NP-hard problems
  (e.g., \cite{li00}).
 Several formal models for this problem have been proposed:
 (a) binary integer programming problem
 (e.g., \cite{frota07,frota10,hos11}),
 (b) model based on the independent set problem
  \cite{hos11},
 and
 (c) two integer programming formulations using  representatives
 \cite{bah14}.

 Fig. 15 depicts an instance of partition coloring problem
 (graph with ten vertices and four graph parts).
 Here, the resultant colorings are
 (two colors: \(c_{1}\), \(c_{1}\)):

 \(Q^{1}  = < 2(c_{1}), 6(c_{2}), 9(c_{1}), 5(c_{2}) >\),
 ~\(Q^{2}  = < 2(c_{2}), 6(c_{1}), 9(c_{2}), 5(c_{1}) >\).

\begin{center}
\begin{picture}(50,48)
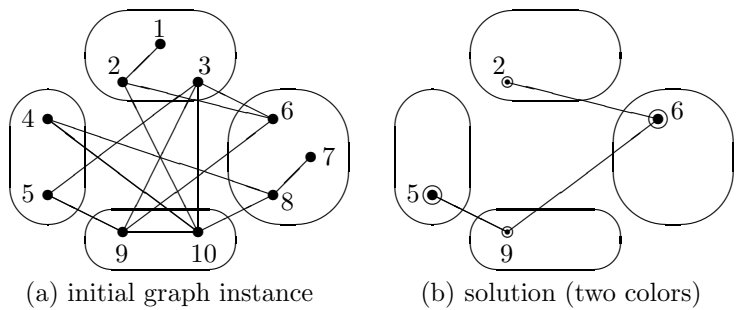
\put(010.5,00){\makebox(0,0)[bl]{Fig. 15. Instance of partition
  coloring problem}}

\put(02,05){\makebox(0,0)[bl]{(a) initial graph instance}}

\put(15,35){\line(1,1){05}}


\put(15,35){\circle*{1.5}} \put(25,35){\circle*{1.5}}
\put(20,40){\circle*{1.5}}

\put(19,41){\makebox(0,0)[bl]{\(1\) }}
\put(13,36){\makebox(0,0)[bl]{\(2\) }}
\put(25,36){\makebox(0,0)[bl]{\(3\) }}

\put(20,38.5){\oval(20,12)}


\put(25,15){\line(0,1){20}} \put(15,15){\line(1,2){10}}
\put(25,15){\line(-1,2){10}}


\put(15,15){\line(1,0){10}}

\put(15,15){\circle*{1.5}} \put(25,15){\circle*{1.5}}

\put(14,11){\makebox(0,0)[bl]{\(9\) }}
\put(24,11){\makebox(0,0)[bl]{\(10\) }}

\put(20,14){\oval(20,08)}

\put(15,15){\line(-2,1){10}} \put(25,15){\line(-4,3){20}}

\put(05,20){\line(4,3){20}}



\put(05,20){\circle*{1.5}} \put(05,30){\circle*{1.5}}

\put(01.5,29){\makebox(0,0)[bl]{\(4\) }}
\put(01.5,19){\makebox(0,0)[bl]{\(5\) }}

\put(05,25){\oval(10,18)}

\put(25,15){\line(2,1){10}} \put(15,15){\line(4,3){20}}

\put(15,15){\line(-2,1){10}} \put(25,15){\line(-4,3){20}}

\put(05,30){\line(3,-1){30}}

\put(35,30){\line(-2,1){10}} \put(35,30){\line(-4,1){20}}



\put(35,20){\line(1,1){05}}


\put(35,20){\circle*{1.5}} \put(35,30){\circle*{1.5}}
\put(40,25){\circle*{1.5}}

\put(36,30){\makebox(0,0)[bl]{\(6\) }}
\put(41.5,24){\makebox(0,0)[bl]{\(7\) }}
\put(36,18){\makebox(0,0)[bl]{\(8\) }}

\put(37,25){\oval(16,18)}

\end{picture}
%
\begin{picture}(45,48)

\put(03.4,05){\makebox(0,0)[bl]{(b)
 solution (two colors)}}



\put(15,35){\circle*{0.8}} \put(15,35){\circle{1.5}}


\put(13,36){\makebox(0,0)[bl]{\(2\) }}

\put(20,38.5){\oval(20,12)}





\put(15,15){\circle*{0.8}} \put(15,15){\circle{1.5}}


\put(14,11){\makebox(0,0)[bl]{\(9\) }}

\put(20,14){\oval(20,08)}

\put(15,15){\line(-2,1){10}}





\put(05,20){\circle*{1.5}} \put(05,20){\circle{2.3}}


\put(01.5,19){\makebox(0,0)[bl]{\(5\) }}

\put(05,25){\oval(10,18)}


\put(15,15){\line(4,3){20}}

\put(15,15){\line(-2,1){10}}




\put(35,30){\line(-4,1){20}}


\put(35,30){\circle*{1.5}} \put(35,30){\circle{2.3}}


\put(36.7,30){\makebox(0,0)[bl]{\(6\) }}


\put(37,25){\oval(16,18)}

\end{picture}
\end{center}

 Some solving approaches proposed for the partition coloring problem
 are listed in Table 11.

\begin{center}
{\bf Table 11.} Algorithms for partition coloring problem    \\
\begin{tabular}{| c | l| l | }
\hline
 No.  &Approach &Source(s) \\
\hline

 1.&Branch-and-price approach &\cite{bah14,frota07,frota10,hos11}\\

 2.&Tabu search heuristic    &\cite{noro06}\\

 3.& Two-phase heuristic  &\cite{noro06}\\
%
 4.&Engineering heuristics &\cite{li00,liu13}\\

\hline
\end{tabular}
\end{center}

 In real world,  this problem corresponds to  routing and
 wavelength assignment in all-optical networks
 (i.e., computation of alternative routes for the lightpaths,
 followed by the solution of a partition colorings problem in a
 conflict graph)
 (e.g., \cite{li00,liu13,noro06}.

 In fact, the  partition coloring problem
 is very close to representative problems
 (e.g., \cite{bah14}).
 Generally,
 this kind of problems is based on selection of elements
 from graph parts (components)
  (e.g., vertices)
  while taking into account
  compatibility of the selected elements
 (i.e., construction of a clique or quasi-clique).
 In addition, it is possible to examine
 some preference relation(s) over elements for graph part.
 Thus, the problem can be considered as a morphological
 clique problem
 (i.e., hierarchical morphological design or combinatorial synthesis)
 \cite{lev98,lev06,lev15}.

 In the future,
 it may be very interesting to examine
 a new multistage partition coloring problem
  with costs of changes of vertex colors
  as restructuring of partition coloring problem.
 (i.e., a version of dynamical partition coloring problem).

\newpage
\section{Some applications}

\subsection{Composite planning framework in paper production system}

 Here a composite planning framework is described
 that was prepared by the author for a seminar of Institute for Industrial Mathematics
 in May 1992 (Beer Sheva, Israel).
 Fig. 16 depicts an illustrative solution of the composite planning problem
 for three machines.

\begin{center}
\begin{picture}(120,100)
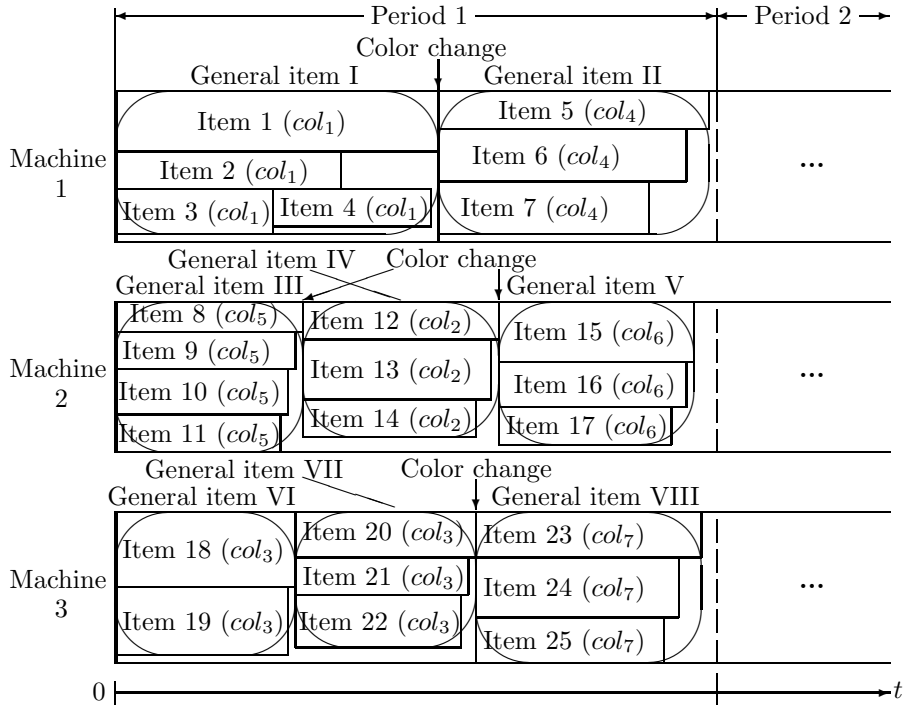
\put(14,00){\makebox(0,0)[bl]{Fig. 16. Composite planning for paper production system}}

\put(14,06){\vector(1,0){103}}

\put(14,4.5){\line(0,1){3}}

\put(94,4.5){\line(0,1){4}}

\put(94,10){\line(0,1){3}} \put(94,14){\line(0,1){3}}
\put(94,18){\line(0,1){3}} \put(94,22){\line(0,1){3}}
\put(94,26){\line(0,1){3}}

\put(94,38){\line(0,1){3}} \put(94,42){\line(0,1){3}}
\put(94,46){\line(0,1){3}} \put(94,50){\line(0,1){3}}
\put(94,54){\line(0,1){3}}

\put(94,66){\line(0,1){3}}
\put(94,70){\line(0,1){3}} \put(94,74){\line(0,1){3}}
\put(94,78){\line(0,1){3}} \put(94,82){\line(0,1){3}}


\put(11,05){\makebox(0,0)[bl]{\(0\)}}

\put(117.5,05.3){\makebox(0,0)[bl]{\(t\)}}

\put(48,095){\makebox(0,0)[bl]{Period \(1\)}}

\put(47,96){\vector(-1,0){33}} \put(62,96){\vector(1,0){32}}
\put(14,97){\line(0,-1){12}} \put(94,97){\line(0,-1){12}}


\put(99,095){\makebox(0,0)[bl]{Period \(2\)}}
\put(98,96){\vector(-1,0){04}}
\put(113,96){\vector(1,0){04}}
\put(00,76){\makebox(0,0)[bl]{Machine}}
\put(06,72){\makebox(0,0)[bl]{\(1\)}}

\put(14,66){\line(0,1){20}}\put(14.31,66){\line(0,1){20}}

\put(14,66){\line(1,0){103}}
\put(14,86){\line(1,0){103}}
\put(57,86){\line(0,-1){20}}
\put(46,90){\makebox(0,0)[bl]{Color change}}
\put(57,90.3){\vector(0,-1){04}}


\put(24,87){\makebox(0,0)[bl]{General item I}}
\put(35.5,76.5){\oval(43,19)}

\put(14,78){\line(1,0){43}} \put(14,86){\line(1,0){43}}
\put(14,78){\line(0,1){08}} \put(57,78){\line(0,1){08}}

\put(25,80){\makebox(0,0)[bl]{Item 1 (\(col_{1}\))}}

\put(14,73){\line(1,0){30}} \put(14,78){\line(1,0){30}}
\put(14,73){\line(0,1){05}} \put(44,73){\line(0,1){05}}

\put(20,73.5){\makebox(0,0)[bl]{Item 2 (\(col_{1}\))}}

\put(14,67){\line(1,0){21}} \put(14,73){\line(1,0){21}}
\put(14,67){\line(0,1){06}} \put(35,67){\line(0,1){06}}

\put(15,68){\makebox(0,0)[bl]{Item 3 (\(col_{1}\))}}

\put(35,68){\line(1,0){21}} \put(35,73){\line(1,0){21}}
\put(35,68){\line(0,1){05}} \put(56,68){\line(0,1){05}}

\put(36,68.5){\makebox(0,0)[bl]{Item 4 (\(col_{1}\))}}


\put(63,87){\makebox(0,0)[bl]{General item II}}
\put(75,76.5){\oval(36,19)}

\put(57,81){\line(1,0){36}} \put(57,86){\line(1,0){36}}
\put(57,81){\line(0,1){05}} \put(93,81){\line(0,1){05}}

\put(65,81.5){\makebox(0,0)[bl]{Item 5 (\(col_{4}\))}}

\put(57,74){\line(1,0){33}} \put(57,81){\line(1,0){33}}
\put(57,74){\line(0,1){07}} \put(90,74){\line(0,1){07}}

\put(61.5,75.5){\makebox(0,0)[bl]{Item 6 (\(col_{4}\))}}

\put(57,67){\line(1,0){28}} \put(57,74){\line(1,0){28}}
\put(57,67){\line(0,1){07}} \put(85,67){\line(0,1){07}}

\put(60,68.5){\makebox(0,0)[bl]{Item 7 (\(col_{4}\))}}


\put(105,76){\makebox(0,0)[bl]{{\bf ...}}}

\put(00,48){\makebox(0,0)[bl]{Machine}}
\put(06,44){\makebox(0,0)[bl]{\(2\)}}

\put(14,38){\line(0,1){20}}\put(14.31,38){\line(0,1){20}}

\put(14,38){\line(1,0){103}}
\put(14,58){\line(1,0){103}}

\put(105,48){\makebox(0,0)[bl]{{\bf ...}}}

\put(50,62){\makebox(0,0)[bl]{Color change}}
\put(65,62.3){\vector(0,-1){04}}
\put(51,62.3){\vector(-3,-1){12}}

\put(14,59){\makebox(0,0)[bl]{General item III}}
\put(26.5,48){\oval(25,20)}

\put(14,54){\line(1,0){25}} \put(14,58){\line(1,0){25}}
\put(14,54){\line(0,1){04}} \put(39,54){\line(0,1){04}}

\put(16,54.5){\makebox(0,0)[bl]{Item 8 (\(col_{5}\))}}

\put(14,49){\line(1,0){24}} \put(14,54){\line(1,0){24}}
\put(14,49){\line(0,1){05}} \put(38,49){\line(0,1){05}}

\put(15,49.5){\makebox(0,0)[bl]{Item 9 (\(col_{5}\))}}

\put(14,43){\line(1,0){23}} \put(14,49){\line(1,0){23}}
\put(14,43){\line(0,1){06}} \put(37,43){\line(0,1){06}}

\put(14.5,44){\makebox(0,0)[bl]{Item 10 (\(col_{5}\))}}
\put(14,38){\line(1,0){22}} \put(14,43){\line(1,0){22}}
\put(14,38){\line(0,1){05}} \put(36,38){\line(0,1){05}}

\put(14.5,38.5){\makebox(0,0)[bl]{Item 11 (\(col_{5}\))}}

\put(21,62.5){\makebox(0,0)[bl]{General item IV}}
\put(40,62.2){\line(3,-1){12}}

\put(52,49){\oval(26,18)}

\put(39,53){\line(1,0){26}} \put(39,58){\line(1,0){26}}
\put(39,53){\line(0,1){05}} \put(65,53){\line(0,1){05}}

\put(40,53.5){\makebox(0,0)[bl]{Item 12 (\(col_{2}\))}}

\put(39,45){\line(1,0){25}} \put(39,53){\line(1,0){25}}
\put(39,45){\line(0,1){08}} \put(64,45){\line(0,1){08}}

\put(40,47){\makebox(0,0)[bl]{Item 13 (\(col_{2}\))}}

\put(39,40){\line(1,0){23}} \put(39,45){\line(1,0){23}}
\put(39,40){\line(0,1){05}} \put(62,40){\line(0,1){05}}

\put(40,40.5){\makebox(0,0)[bl]{Item 14 (\(col_{2}\))}}

\put(66,59){\makebox(0,0)[bl]{General item V}}
\put(78,48.5){\oval(26,19)}

\put(65,50){\line(1,0){26}} \put(65,58){\line(1,0){26}}
\put(65,50){\line(0,1){08}} \put(91,50){\line(0,1){08}}

\put(67,52){\makebox(0,0)[bl]{Item 15 (\(col_{6}\))}}

\put(65,44){\line(1,0){25}} \put(65,50){\line(1,0){25}}
\put(65,44){\line(0,1){06}} \put(90,44){\line(0,1){06}}

\put(67,45){\makebox(0,0)[bl]{Item 16 (\(col_{6}\))}}

\put(65,39){\line(1,0){23}} \put(65,44){\line(1,0){23}}
\put(65,39){\line(0,1){05}} \put(88,39){\line(0,1){05}}

\put(66,39.5){\makebox(0,0)[bl]{Item 17 (\(col_{6}\))}}

\put(00,20){\makebox(0,0)[bl]{Machine}}
\put(06,16){\makebox(0,0)[bl]{\(3\)}}

\put(14,10){\line(0,1){20}} \put(14.31,10){\line(0,1){20}}

\put(14,10){\line(1,0){103}}
\put(14,30){\line(1,0){103}}


\put(105,20){\makebox(0,0)[bl]{{\bf ...}}}

\put(52,34){\makebox(0,0)[bl]{Color change}}
\put(62,34.3){\vector(0,-1){04}}

\put(13,31){\makebox(0,0)[bl]{General item VI}}
\put(26,20.5){\oval(24,19)}

\put(14,20){\line(1,0){24}} \put(14,30){\line(1,0){24}}
\put(14,20){\line(0,1){10}} \put(38,20){\line(0,1){10}}

\put(15,23){\makebox(0,0)[bl]{Item 18 (\(col_{3}\))}}

\put(14,11){\line(1,0){23}} \put(14,20){\line(1,0){23}}
\put(14,11){\line(0,1){09}} \put(37,11){\line(0,1){09}}

\put(15,13.5){\makebox(0,0)[bl]{Item 19 (\(col_{3}\))}}

\put(18,34.5){\makebox(0,0)[bl]{General item VII}}
\put(39,34.7){\line(3,-1){12.2}}

\put(50,21){\oval(24,18)}

\put(38,24){\line(1,0){24}} \put(38,30){\line(1,0){24}}
\put(38,24){\line(0,1){06}} \put(62,24){\line(0,1){06}}

\put(39,25.5){\makebox(0,0)[bl]{Item 20 (\(col_{3}\))}}

\put(38,19){\line(1,0){23}} \put(38,24){\line(1,0){23}}
\put(38,19){\line(0,1){05}} \put(61,19){\line(0,1){05}}

\put(39,19.5){\makebox(0,0)[bl]{Item 21 (\(col_{3}\))}}


\put(38,12){\line(1,0){22}} \put(38,19){\line(1,0){22}}
\put(38,12){\line(0,1){07}} \put(60,12){\line(0,1){07}}

\put(38.5,13.5){\makebox(0,0)[bl]{Item 22 (\(col_{3}\))}}

\put(64,31){\makebox(0,0)[bl]{General item VIII}}
\put(77,20){\oval(30,20)}

\put(62,24){\line(1,0){30}} \put(62,30){\line(1,0){30}}
\put(62,24){\line(0,1){06}} \put(92,24){\line(0,1){06}}

\put(63,25){\makebox(0,0)[bl]{Item 23 (\(col_{7}\))}}

\put(62,16){\line(1,0){27}} \put(62,24){\line(1,0){27}}
\put(62,16){\line(0,1){08}} \put(89,16){\line(0,1){08}}

\put(63,18){\makebox(0,0)[bl]{Item 24 (\(col_{7}\))}}


\put(62,10){\line(1,0){25}} \put(62,16){\line(1,0){25}}
\put(62,10){\line(0,1){06}} \put(87,10){\line(0,1){06}}

\put(63,11){\makebox(0,0)[bl]{Item 25 (\(col_{7}\))}}

\end{picture}
\end{center}
 In the problem, there are a set of paper horizontal bar for each machine.
 It is necessary to cut it (by special knifes) to obtain a set of 2D items of the required sizes and colors
 (by coloring).
 Seven colors are considered:
 white (\(col_{1}\)),
 blue (\(col_{2}\)),
 red (\(col_{3}\)),
 green (\(col_{4}\)),
 magenta (\(col_{5}\)),
 brown (\(col_{6}\)), and
 yellow (\(col_{7}\)).
 Table 12 contains
 ordinal estimates of color change:  \(col_{i} \Rightarrow col_{j}\)
 (\(i= \overline{1,7}\), \(j = \overline{1,7}\)).
 Item parameters are presented in Table 13:
 twenty five 2D items (required items of required sizes and colors)
 (the width of the paper horizontal bar equals \(20\)).

  Evidently,
  two objective functions are considered:

 (i) minimizing the volume of non-used domain in bins,

 (ii) minimizing the total cost of color changes
 (e.g., as a total sum of color change estimates in the solution)
 (this function can be transformed to non-used bin domains as well).

\begin{center}
  Table 12. Ordinal estimates of color change
  (\(col_{i} \Rightarrow col_{j}\)) \\
\begin{tabular}{  |l|c|c|c| c|c|c|c | }
\hline
   \(col_{i} \backslash col_{j}\)
  &\(col_{1}\) &\(col_{2}\) &\(col_{3}\) &\(col_{4}\) &\(col_{5}\) &\(col_{6}\) &\(col_{7}\)
   \\
\hline

 \(col_{1}\) (white)  &\(0\)&\(0\)&\(0\)&\(0\)&\(0\)&\(0\)&\(0\)   \\
 \(col_{2}\) (blue)   &\(4\)&\(0\)&\(4\)&\(4\)&\(2\)&\(1\)&\(3\)   \\
 \(col_{3}\) (red)    &\(4\)&\(0\)&\(0\)&\(4\)&\(3\)&\(0\)&\(3\)   \\
 \(col_{4}\) (green)  &\(4\)&\(4\)&\(4\)&\(0\)&\(3\)&\(0\)&\(5\)   \\
 \(col_{5}\) (magenta)&\(4\)&\(0\)&\(3\)&\(4\)&\(0\)&\(0\)&\(3\)   \\
 \(col_{6}\) (brown)  &\(4\)&\(4\)&\(4\)&\(4\)&\(4\)&\(0\)&\(4\)   \\
 \(col_{7}\) (yellow) &\(2\)&\(0\)&\(2\)&\(3\)&\(1\)&\(0\)&\(0\)   \\

\hline
\end{tabular}
\end{center}

 The following heuristic solving scheme is considered:

~~

 {\it Stage 1.} Grouping of initial items by colors.

 {\it Stage 2.} For each color:
 forming the general items
  (combinations of items of the same color)
  as packed bins
 (bin size equals  \(20\)).
 For the items in the same bin,
 their heights/lenghts
  are about close.
 Some initial items can be integrated
 (as items 3 and 4 in the example, Fig. 16).
 Here, bin packing problem can be used.
 As e result, a set of general items
 (the same color for each item)
 are obtained.
 In Fig. 16, the following \(8\) general items
 are depicted:
 (i) items \(1\), \(2\), and \(3 \& 4\) (color \(col_{1}\));
 (ii) items \(5\), \(6\), and \(7\) (color \(col_{4}\));
 (iii) items \(8\), \(9\), \(10\), and \(11\) (color \(col_{5}\));
 (iv) items \(12\), \(13\), and \(14\) (color \(col_{2}\));
 (v) items \(15\), \(16\), and \(17\) (color \(col_{6}\));
 (vi) items \(18\) and \(19\) (color \(col_{3}\));
 (vii) items \(20\), \(21\), and \(22\) (color \(col_{3}\));
 and
 (viii) items \(23\), \(24\), and \(25\) (color \(col_{7}\)).

 {\it Stage 3.} Forming  the bins
 for each machine and for one period
 (from the general items):
 bin size corresponds to time period).
 Here bin packing problem can be used.

 {\it Stage 4.}
 For each obtained bin:
 linear ordering of the generalized items
 while taking into account
 color changes.
 Here the traveling salesman problem can be used
 (while taking into account the ordinal estimates of color change
 as element distance, Table 12).

~~

\begin{center}
  Table 13. Items and their parameters  \\
\begin{tabular}{  |c|c|c|c| c|c|c| }
\hline
 Item&Width &Height/lenght &Color &General item &Machine &Time interval \\
\hline

 \(1\)   &\(8\)&\(43\)&\(col_{1}\)&\(I\)&\(1\)&\(1\)   \\
 \(2\)   &\(5\)&\(30\)&\(col_{1}\)&\(I\)&\(1\)&\(1\)   \\
 \(3\)   &\(6\)&\(21\)&\(col_{1}\)&\(I\)&\(1\)&\(1\)   \\
 \(4\)   &\(5\)&\(21\)&\(col_{1}\)&\(I\)&\(1\)&\(1\)   \\

 \(5\)   &\(5\)&\(36\)&\(col_{4}\)&\(II\)&\(1\)&\(2\)   \\
 \(6\)   &\(7\)&\(33\)&\(col_{4}\)&\(II\)&\(1\)&\(2\)   \\
 \(7\)   &\(7\)&\(28\)&\(col_{4}\)&\(II\)&\(1\)&\(2\)   \\

 \(8\)   &\(4\)&\(25\)&\(col_{5}\)&\(III\)&\(2\)&\(1\)   \\
 \(9\)   &\(5\)&\(24\)&\(col_{5}\)&\(III\)&\(2\)&\(1\)   \\
 \(10\)  &\(6\)&\(23\)&\(col_{5}\)&\(III\)&\(2\)&\(1\)   \\
 \(11\)  &\(5\)&\(22\)&\(col_{5}\)&\(III\)&\(2\)&\(1\)   \\

 \(12\)  &\(5\)&\(26\)&\(col_{2}\)&\(IV\)&\(2\)&\(2\)   \\
 \(13\)  &\(8\)&\(25\)&\(col_{2}\)&\(IV\)&\(2\)&\(2\)   \\
 \(14\)  &\(5\)&\(23\)&\(col_{2}\)&\(IV\)&\(2\)&\(2\)   \\

 \(15\)  &\(8\)&\(26\)&\(col_{6}\)&\(V\)&\(2\)&\(3\)   \\
 \(16\)  &\(6\)&\(25\)&\(col_{6}\)&\(V\)&\(2\)&\(3\)   \\
 \(17\)  &\(5\)&\(23\)&\(col_{6}\)&\(V\)&\(2\)&\(3\)   \\

 \(18\) &\(10\)&\(24\)&\(col_{3}\)&\(VI\)&\(3\)&\(1\)   \\
 \(19\)  &\(9\)&\(23\)&\(col_{3}\)&\(VI\)&\(3\)&\(1\)   \\

 \(20\)  &\(6\)&\(24\)&\(col_{3}\)&\(VII\)&\(3\)&\(2\)   \\
 \(21\)  &\(5\)&\(23\)&\(col_{3}\)&\(VII\)&\(3\)&\(2\)   \\
 \(22\)  &\(7\)&\(22\)&\(col_{3}\)&\(VII\)&\(3\)&\(2\)   \\

 \(23\)  &\(6\)&\(30\)&\(col_{7}\)&\(VIII\)&\(3\)&\(3\)   \\
 \(24\)  &\(8\)&\(27\)&\(col_{7}\)&\(VIII\)&\(3\)&\(3\)   \\
 \(25\)  &\(6\)&\(25\)&\(col_{7}\)&\(VIII\)&\(3\)&\(3\)   \\

\hline
\end{tabular}
\end{center}

 Note the considered composite planning framework can be extended/modifued to use
 in communication systems (e.g., multiple channel systems).

\subsection{Planning in communication system}

 The basic multi-processor scheduling problems
 based on bin packing have been described
 in
 \cite{gar87,cof79,conw67,hub94}.
 Here,
 some combinatorial planning problems
 as 2D bin packing
 for communications
 (one-channel communications,
 telecommunication WiMAX systems).
 Note, close
 problems are used in resource allocation in multispot satellite networks
 (e.g., \cite{alouf06}).

\subsubsection{Selection of messages/information packages}

 First, the basic simplified planning problem can be considered as
 the well-known secretary problem.
 Given a set of items \(n\) (e.g., messages)
 \(A = \{a_{1},...,a_{i},...,a_{n}\}\),
 each item \(a_{i}\) has a weight \(w_{i}\)
 (e.g., time for processing).
 The problem is (Fig. 17):

~~

 Find the schedule
 (i.e., ordering of items as  permutation)
 of the items from \(A\):

 \( S = < s[1],...,s[\iota],...,s[n] > \)
 (\(s[\iota]\)) corresponds to an item \(a_{i}\) that is processed
 at the \(\iota\)-th place in schedule \(S\))
 such that average  completion time
 for each item \(a_{i} \in A\)
 (i.e., sum of waiting time and processing time)
 is minimal:
 ~\( t(S) = \frac{1}{n} \sum_{\iota =1}^n ~ \tau_{s[\iota]} \),
 where the waiting time is as follows
  (\(\tau_{s[1]} = w_{s[1]}\), ~\(\iota = \overline{2,n}\)):~
 \(\tau_{s[\iota]} = w_{s[\iota]} + \sum_{\kappa =1}^{\iota - 1}  w_{s[\kappa]}  \)
 ~\(\forall \kappa=\overline{1,n} \).

~~

 Evidently, the algorithm to obtain the optimal solution is based on ordering of the items by
 non-decreasing of \(w_{i}\)
 (i.e., the item with minimal weight has to be processed as the
 1st, and so on) (complexity estimate of the algorithm is \(O(n \log n)\)).
 This is the algorithm: 'smallest weight first'.

 Note,
 the solution can be defined by Boolean variables:
 \(x_{a_{i},s[\iota]} \in \{0,1\}\), where \(x_{a_{i},s[\iota]} =1\)
 if item \(a_{i}\) is assigned
 into place \(s[\iota]\) in the solution.
 Thus, the solution is defined by Boolean matrix:~

 \( X = || x_{a_{i},s[\iota]}  || \),
  \(i=\overline{1,n}\),
  \(\iota=\overline{1,n}\).

\begin{center}
\begin{picture}(91,24)
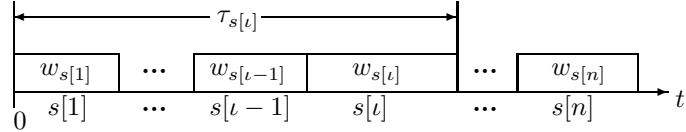
\put(011,00){\makebox(0,0)[bl]{Fig. 17. Illustration for
  secretary problem}}

\put(00,10){\vector(1,0){87}} \put(00,08){\line(0,1){04}}

\put(00,05){\makebox(0,0)[bl]{\(0\)}}
\put(88,08){\makebox(0,0)[bl]{\(t\)}}

\put(04.5,06){\makebox(0,0)[bl]{\(s[1]\)}}

\put(03.5,11){\makebox(0,0)[bl]{\(w_{s[1]}\)}}

\put(00,10){\line(0,1){05}} \put(14,10){\line(0,1){05}}

\put(00,15){\line(1,0){14}}


\put(17,07.5){\makebox(0,0)[bl]{{\bf ...}}}
\put(17,12.5){\makebox(0,0)[bl]{{\bf ...}}}

\put(26,06){\makebox(0,0)[bl]{\(s[\iota-1]\)}}

\put(26,11){\makebox(0,0)[bl]{\(w_{s[\iota- 1]}\)}}

\put(24,10){\line(0,1){05}} \put(39,10){\line(0,1){05}}

\put(24,15){\line(1,0){15}}

\put(45,06){\makebox(0,0)[bl]{\(s[\iota]\)}}

\put(45,11){\makebox(0,0)[bl]{\(w_{s[\iota]}\)}}

\put(39,10){\line(0,1){05}} \put(59,10){\line(0,1){05}}

\put(39,15){\line(1,0){20}}


\put(61,07.5){\makebox(0,0)[bl]{{\bf ...}}}
\put(61,12.5){\makebox(0,0)[bl]{{\bf ...}}}

\put(71.5,06){\makebox(0,0)[bl]{\(s[n]\)}}

\put(71.5,11){\makebox(0,0)[bl]{\(w_{s[n]}\)}}

\put(67,10){\line(0,1){05}} \put(83,10){\line(0,1){05}}

\put(67,15){\line(1,0){16}}


\put(59,10){\line(0,1){12}} \put(00,10){\line(0,1){12}}

\put(26,20){\vector(-1,0){26}} \put(33,20){\vector(1,0){26}}

\put(27,18){\makebox(0,0)[bl]{\(\tau_{s[\iota]}\)}}

\end{picture}
\end{center}

 Usually  the described secretary problem is used
 for planning in one-channel communication system.
 In this case, there is a time interval
 (i.e., planning period \(T\))
 and the initial set of items \(A\)
 is ordered to send via the channel.
 If all messages can be send during  period \(T\)
 (i.e., \( \sum_{i=1}^{n} w_{i}  \leq T  \))
 the considered algorithm can be successfully used.
 Unfortunately,
 if period \(T\) is not sufficient to send all message
(i.e., \( \sum_{i=1}^{n} w_{i}  \geq T  \)),
 a subset of items with highest weights
 have to wait the next period (i.e., a wait set).
 Here, the problem can be formulated as a knapsack model:
 \[\min~ t(S) = \frac{1}{n} \sum_{\iota =1}^n ~ \tau_{s[\iota]}  ~x_{a_{i},s[\iota]},
 ~~~~~ \max~  \sum_{i=1}^{n} \sum_{\iota=1}^{n}   ~x_{a_{i},s[\iota]}
 \]
 \[s.t. ~~~ \sum_{i=1}^{n} \sum_{\iota=1}^{n} ~x_{a_{i},s[\iota]}
~w_{i}
   \leq T,
%
 ~~~ \sum_{\iota}^{n} ~x_{a_{i},s[\iota]} \leq 1 ~~ \forall i=\overline{1,n},
%
 ~~~ x_{a_{i},s[\iota]} \in \{0,1\}.\]
 Here, the algorithm above leads to the optimal solution.
 Note, the first objective function requires
 linear ordering of items in the solution
 by non-decreasing of \(w_{i}\)
 (as in previous problem).

 After using the algorithm
 the items which do not belong to the solution
 can be considered as a wait set.
 Thus, it is reasonable to examine an extension of the
 problem above.
 Let each item \(a_{i} \in A\) has two parameters:
 (i) the weight (i.e., processing time) \(w_{i}\) and
 (ii) the number of wait periods \(\gamma_{i} = 0,1,...\) ~.
 The problem statement can be considered as two-criteria knapsack model:
 \[\min~ t(S) = \frac{1}{n}  \sum_{\iota =1}^n ~ \tau_{s[\iota]}  ~x_{a_{i},s[\iota]},
%
 ~~~~~ \max~  \sum_{i=1}^{n} \sum_{\iota=1}^{n}   ~x_{a_{i},s[\iota]},
 ~~~~~ \max~  \sum_{i=1}^{n} \sum_{\iota=1}^{n}
   ~x_{a_{i},s[\iota]} ~\gamma_{i} \]
\[s.t. ~~~ \sum_{i=1}^{n} \sum_{\iota=1}^{n} ~x_{a_{i},s[\iota]}
 ~w_{i}  \leq T,
%
 ~~~ x_{a_{i},s[\iota]} \in \{0,1\}.\]
 This problem is NP-hard.
 The selection of items for sending
 (i.e., solution)
 can be based on
 detection of Pareto-efficient items by two parameters:
 ~(a) minimum weight \(w_{i}\)
 (rule: smallest weight first) and
 ~(b) maximum number \(\gamma_{i}\)
 (rule: longest wait first).
%
%
 The following heuristic algorithm can be considered:

~~

 {\it Stage 1.} Definition \(\widehat{A} = A\).

 {\it Stage 2.}
 Deletion  of Pareto-efficient items
 in \(\widehat{A}\)
 by two parameters
 weight \(w_{i}\) (minimum) and
 importance  \(\gamma_{i}\) (maximum)
 to obtain the subset \(A^{P}\subseteq  \widehat{A}\)
 (the current items layer by Pareto rule).

 {\it Stage 3.} Assignment of items from \(A^{P}\) to bins.

 {\it Stage 4.}
 Definition subset \( \widehat{A}= A \backslash A^{P} \).
 If \( |\widehat{A} | = 0\) that GO TO Stage 5
 Otherwise GO TO Stage 2.

 {\it Stage 5.} Stop.

~~

  Complexity estimates for the above-mentioned version hierarchical clustering algorithm (by stages)
   is presented in Table 14.

\begin{center}
{\bf Table 14.}  Complexity estimates\\
\begin{tabular}{| l | l|c |}
\hline
  Stage &Description&Complexity estimate\\
   & &(running time)\\

\hline

 Stage 1 &  Definition \(\widehat{A} = A\). &\(O (1)\) \\

 Stage 2 &
 Deletion  of current Pareto-efficient items layer
 \(A^{P}\subseteq  \widehat{A}\) in \(\widehat{A}\)
  &\(O (n^{2})\) \\

 & (by parameters \(w_{i}\) and \(\gamma_{i}\))&\\

 Stage 3.& Assignment of items from \(A^{P}\) to bins. &\(O(n)\)\\

 Stage 4 & \( \widehat{A}= A \backslash A^{P} \). If all items are
 processed GO TO Stage 2.
 &\(O (1)\) \\
  & Otherwise GO TO Stage 5. &\\

 Stage 5.& Stopping         & \(O(1)\)\\

\hline
\end{tabular}
\end{center}


 Afterhere,
 the first objective function
 \(\min~ t(S) = \frac{1}{n} \sum_{\iota =1}^n ~ \tau_{s[\iota]}
  ~x_{a_{i},s[\iota]}\)
 will not be considered
 because items of the solution
 can be ordered to take into account n the objective function.

 Evidently,
 each item (message) can have other parameters, for example,
 importance
 (it will leads to an additional objective function in the model above).
 In this case, the model is:
 \[\max~ \sum_{i=1}^{n}   \sum_{\iota =1}^n ~ \beta_{a_{i}}  ~x_{a_{i},s[\iota]},
%
 ~~~~  \max~ \sum_{i=1}^{n} \sum_{\iota=1}^{n}   ~x_{a_{i},s[\iota]},
 ~~~~  \max~ \sum_{i=1}^{n} \sum_{\iota=1}^{n}
   ~x_{a_{i},s[\iota]} ~\gamma_{i} \]
\[s.t. ~~~ \sum_{i=1}^{n} \sum_{\iota=1}^{n} ~x_{a_{i},s[\iota]}
 ~w_{i} \leq T,
%
 ~~~ x_{a_{i},s[\iota]} \in \{0,1\},\]
 where
 \(\beta_{a_{i}}\) is importance parameter of the corresponding item \(i\).
 Note the importance parameter may by dependent on scheduling place
 \(s[\iota]\):
 ~\(\beta_{a_{i},s[\iota]}\).

 In the case of multiset estimate of the importance parameter
 ~\(e_{a_{i},s[\iota]}\),
 the model is:
%
%
%
 \[ \max~ M = arg \min_{M\in D} ~~
 | \biguplus_{ i \in \{ i | x_{a_{i},s[\iota]}=1   \}} ~   \delta (M,e_{i})|,
%
%
 ~~~~~ \max~ \sum_{i=1}^{n} \sum_{\iota=1}^{n}   ~x_{a_{i},s[\iota]},
 ~~~~~ \max~ \sum_{i=1}^{n} \sum_{\iota=1}^{n}
   ~x_{a_{i},s[\iota]} ~\gamma_{i} \]
\[s.t. ~~~ \sum_{i=1}^{n} \sum_{\iota=1}^{n} ~x_{a_{i},s[\iota]}
 ~w_{i} \leq T,
%
 ~~~ x_{a_{i},s[\iota]} \in \{0,1\},\]
 In addition,
 precedence binary relation over items can be examined as well.
 This leads to an additional logical constraint in the model above
 and corresponding algorithm scheme
 is based on linear ordering
 of the selected items
 while taking account the precedence constraint.

\subsubsection{Two-dimensional packing in WiMAX system}

 In recent decade,
 two-dimensional packing problems have been used
 in contemporary  telecommunication systems
 (IEEE 802.16/WiMAX standard)
  \cite{cic10,cic11,cic14,lodi11,mar14}.

 An illustrative structure of WiMAX system is depicted in Fig. 18.

\begin{center}
\begin{picture}(70,56)
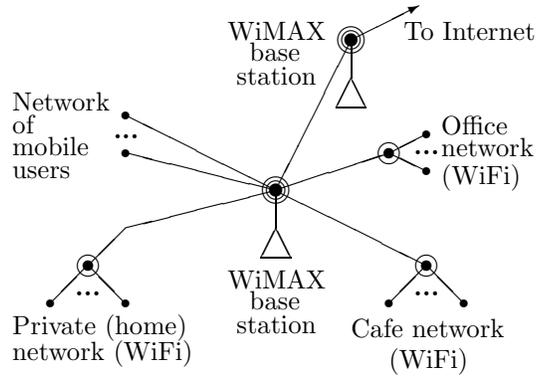
\put(04.5,00){\makebox(0,0)[bl]{Fig. 18. Structure of WiMAX system}}


\put(28.6,50){\makebox(0,0)[bl]{WiMAX}}
\put(31.5,47.2){\makebox(0,0)[bl]{base}}
\put(29.8,44){\makebox(0,0)[bl]{station}}

\put(45,50){\circle*{1.7}} \put(45,50){\circle{2.5}}
\put(45,50){\circle{3.5}}

\put(45,50){\line(0,-1){5}}

\put(45,50){\vector(2,1){09}}
\put(52,50){\makebox(0,0)[bl]{To Internet}}

\put(45,50){\line(-1,-2){10}}

\put(43,41){\line(1,0){4}} \put(43,41){\line(1,2){2}}
\put(47,41){\line(-1,2){2}}


\put(28.6,17){\makebox(0,0)[bl]{WiMAX}}
\put(31.5,14.2){\makebox(0,0)[bl]{base}}
\put(29.8,11){\makebox(0,0)[bl]{station}}


\put(35,30){\circle*{1.7}} \put(35,30){\circle{2.5}}
\put(35,30){\circle{3.5}}

\put(35,30){\line(0,-1){5}}

\put(33,21){\line(1,0){4}} \put(33,21){\line(1,2){2}}
\put(37,21){\line(-1,2){2}}


\put(35,30){\line(3,1){15}}

\put(35,30){\line(-4,1){20}} \put(35,30){\line(-2,1){20}}


\put(35,30){\line(-4,-1){20}} \put(15,25){\line(-1,-1){05}}

\put(35,30){\line(2,-1){20}}


\put(57,37.5){\makebox(0,0)[bl]{Office}}
\put(57,34.5){\makebox(0,0)[bl]{network}}
\put(57,30){\makebox(0,0)[bl]{(WiFi)}}

\put(50,35){\circle*{1.5}} \put(50,35){\circle{2.7}}

\put(50,35){\line(2,1){05}} \put(50,35){\line(2,-1){05}}

\put(55,37.5){\circle*{1.0}}

\put(55,32.5){\circle*{1.0}}

\put(53.3,34.8){\makebox(0,0)[bl]{{\bf ...}}}

\put(45,10.5){\makebox(0,0)[bl]{Cafe network}}
\put(50,05.5){\makebox(0,0)[bl]{(WiFi)}}

\put(55,20){\circle*{1.5}} \put(55,20){\circle{2.7}}

\put(55,20){\line(-1,-1){05}} \put(55,20){\line(1,-1){05}}
\put(50,15){\circle*{1.0}} \put(60,15){\circle*{1.0}}

\put(53.3,16){\makebox(0,0)[bl]{{\bf ...}}}

\put(00,40.6){\makebox(0,0)[bl]{Network}}
\put(00,37.6){\makebox(0,0)[bl]{of}}
\put(00,34.6){\makebox(0,0)[bl]{mobile}}
\put(00,31.6){\makebox(0,0)[bl]{users}}


\put(15,35){\circle*{1.0}} \put(15,40){\circle*{1.0}}

\put(13.4,37){\makebox(0,0)[bl]{{\bf ...}}}

\put(00,10){\makebox(0,0)[bl]{Private (home)}}
\put(00,06.5){\makebox(0,0)[bl]{network (WiFi)}}

\put(10,20){\circle*{1.5}} \put(10,20){\circle{2.7}}

\put(10,20){\line(-1,-1){05}} \put(10,20){\line(1,-1){05}}
\put(05,15){\circle*{1.0}}\put(15,15){\circle*{1.0}}

\put(08.3,16){\makebox(0,0)[bl]{{\bf ...}}}

\end{picture}
\end{center}

 A general description of the above-mentioned approach
 is presented in \cite{mar14} as follows.
 Information transmission process
 is based on rectangular frames
  ``down link zones'':
 time (width)  \(\times\) frequency (height).
 Thus, data packages correspond to 2D items (i.e., rectangular)
 which are stored in ``down link zones'' (i.e., bins).
 In \cite{mar14},
 a general three phase solving scheme is examined:

~~

 {\it Phase 1.} Selection of information packages
 (messages)
 for the current transmission period.

 {\it Phase 2.} Arranging the selected packets into rectangular regions
 (as general items).

 {\it Phase 3.} Allocation of the resultant regions
 to the rectangular frame.

~~

 Note, the above-mentioned phase 1
 can be based on model and solving approach from the previous section
 as selection of Pareto-efficient messages
 (information packages)
  for the current transmission period.
 The allocation problem above (i.e., phase 3) is studied in
  \cite{cic10,cic11,cic14}
  (including problem statement, complexity, heuristic algorithms,
  computing experiments).
 In mobile broadband wireless access systems like IEEE 802.16/WiMAX,
   Orthogonal Frequency Division Multiple Access (OFDMA)
 is used in order to exploit frequency and multi-user diversity
 (i.e., improving the spectral efficiency).
 MAC (medium access control)  frame extends in two dimensions,
 i.e., time and frequency.
 At the beginning of each frame, i.e., every \(5\) ms,
 the base station is responsible both for scheduling packets,
 based on the negotiated quality of service requirements, and
 for allocating them into the frame,
 according to the restrictions imposed by 802.16 OFDMA.

 Here, a two-stage solving scheme for resource allocation is applied
  (e.g., \cite{cic10,cic11,cic14}):
 (a) scheduling of packets in a given time frame,
 (b) allocation of packets across different subcarriers and time slots.
 The  second stage above can be examined as
 a special 2D bin packing problem
 \cite{cic10,cic11,cic14}.

 Evidently, integrated solving scheme for the two above-mentioned stages
 is a prospective research  direction.
 In general,
 it is necessary to study the integrated approach for three-phase
 for planning in WiMAX system from \cite{mar14}.
 In addition,
  it may be prospective to examine ordinal and/or multiset estimates
 for problem elements including
 lattice-based quality domain(s) for problem solutions.

\newpage
\section{Conclusion}

 In this paper,
 a generalized integrated glance
 to bin packing problems
 is suggested.
 The approach is based on
 a system structural problem description:
 (a) element sets
 (i.e., item set, bin set, item subset assigned to bin),
 (b) binary relation over the sets above:
 relation over item set(s) as compatibility,
 precedence, dominance;
 relation over items and bins
 (i.e., correspondence of items to bins).
 Here, the following objective functions can be examined:
 (1) traditional functions
 (i.e., minimizing the number of used bins,
 maximizing the number of assigned items),
 (2) weighted and vector versions of
 the functions above,
  and
  (3) the objective functions based on lattices.
 Some new problem statements with multiset estimates of items
 are presented.
 Two applied examples are considered:
 (i) planning in paper industry,
 (ii) planning in communication systems
  (selection of  messages, packing of massages
  in  WiMAX).

 Generally, it is necessary to point out the following.
 In recent decades there exists a trend
 in applied combinatorial optimization
 (e.g., \cite{lev11ADES,lev14,lev15}):

~~

 ``FROM basic combinatorial problem TO
 composite framework consisting of several interconnected combinatorial problems''.

~~

 A well-known example of the composite frame
 is the following:
 timetabling problem that is usually based
 a combination of basic combinatorial optimization problems
 (e.g., assignment, clustering, graph coloring, scheduling).
 From this viewpoint,
 bin packing problems and their
 extensions/modifications can be examined as a basis of
 various applied composite frameworks.
 Thus, our material may be
  useful to
 build the applied composite frameworks above.

 In the future,
 it may be reasonable
  to investigate the following research directions:

 {\it 1.} further examination of
 bin packing problems
 with
  multiset estimates;

%
 {\it 2.} study of various versions of
 colored bin packing problems
 (e.g.,
 various color proximities,
 various objective functions);

 {\it 3.} examination
 of multi-stage bin packing problems
 (i.e., models, methods, applications);

 {\it 4.} examination of new applied composite frameworks
 based on bin packing problems;

 {\it 5.} execution of computing experiments
 to compare many solving schemes for
 various bin packing problems
 with ordinal/multiset estimates;

 {\it 6.} examination
 of multi-period (or cyclic)
 multi-channel scheduling problems
 based on various bin packing models;

 {\it 7.} study of
  resource allocation in multispot satellite networks
  on the basis of various bin packing problems;

 {\it 8.} analysis of  applied
 multicriteria bin packing problems and
 bin packing problems with
  multiset estimates;
 and

 {\it 9.} usage of our material
 in educational courses
 (e.g., applied mathematics, computer
 science, engineering, management).

\newpage

\end{document}